\pdfoutput=1

\documentclass[11pt]{article}

\usepackage[ruled,vlined]{algorithm2e}
\usepackage{amsmath,amssymb,amsfonts}
\usepackage[noend]{algpseudocode}
\usepackage{bm}	
\usepackage{bbm}
\usepackage{graphicx}
\usepackage[final]{acl}
\usepackage{booktabs} 
\usepackage{times}
\usepackage{latexsym}
\usepackage{amsmath}
\usepackage{graphicx}
\usepackage{subcaption}
\usepackage{url}
\usepackage{soul}
\usepackage{hyperref}
\usepackage[T1]{fontenc}

\usepackage[utf8]{inputenc}

\usepackage{microtype}

\newcommand{\redbold}[1]{\textcolor{black}{\text{#1}}}

\definecolor{lightskyblue}{rgb}{0.53, 0.81, 0.98}
\usepackage{amssymb,amsmath,amsthm,enumitem}
\title{Generalizability of Mixture of Domain-Specific Adapters from the Lens of Signed Weight Directions and its Application to Effective Model Pruning}

\author{Tuc Nguyen \\
  Department of Computer Science \\
  Indiana University \\
  \texttt{nguyentuc1003@gmail.com} \\\And
  Thai Le \\
  Department of Computer Science \\
  Indiana University \\
  \texttt{leqthai.vn@gmail.com} \\}

\begin{document}
\captionsetup[figure]{font=small}
\maketitle
\begin{abstract}
Several parameter-efficient fine-tuning methods based on adapters have been proposed as a streamlined approach to incorporate not only a single specialized knowledge into existing Pre-Trained Language Models (PLMs) but also multiple of them at once. Recent works such as AdapterSoup propose to mix not all but only a selective sub-set of domain-specific adapters during inference via model weight averaging to optimize performance on novel, unseen domains with excellent computational efficiency. However, the essential generalizability of this emerging weight-space adapter mixing mechanism on \textit{unseen, in-domain examples} remains unexplored.
Thus, in this study, we conduct a comprehensive analysis to elucidate the generalizability of domain-specific adapter mixtures in in-domain evaluation. We also provide investigations into the inner workings of the mixture of domain-specific adapters by analyzing their weight signs, yielding critical analysis on the negative correlation between their fraction of weight sign difference and their mixtures' generalizability.
The code is available at \href{https://github.com/nguyentuc/mixture_of_domain_adapter/}{Github}.
\end{abstract}

\section{Introduction}
Recently, several \textit{parameter-efficient fine-tuning methods that are based on adapters} have been introduced as a streamlined approach for fine-tuning Pre-trained Language Models (PLMs) to equip them with new, specialized knowledge or domain. 
Several algorithms have been proposed to train a distinct adapter for each new domain~\cite{houlsby2019parameter, pfeiffer2020adapterfusion, hu2021lora}. 
To further improve a model's generalizability, existing works~\cite{pfeiffer2020adapterfusion,wang2020k,diao2023mixture} mostly focus on training multiple adapters for multiple tasks and continuously adding more adapters for incoming new tasks.
This can be inefficient for the new domain tasks that have only a few examples, making the learning among the tasks unequal. 
Thus, more recent works such as \citet{fisheravg, wang-etal-2022-adamix, wang-etal-2021-efficient-test, li2022branch, chronopoulou2023adaptersoup} opt for weight-space averaging of model and/or adapters trained on different domains, resulting in so-called \textit{Mixture of Expert Adapters}.
 
One recent notable work in this space is AdapterSoup~\cite{chronopoulou2023adaptersoup}, which proposes to merge the weights of a fixed-size, selective sub-set of different domain-specific adapters via an averaging function to accommodate unseen tasks or domains during inference. 
Such weight-space merging mechanism on adapters is efficient in practice as one can efficiently train a small, additional adapter and plug it into existing PLMs to incorporate new knowledge.
Although the work reported favorable evaluation results on unseen, novel domains, it is unclear to what extent such weight-space merging mechanism on domain-specific adapters can generalize in an in-domain evaluation setting--i.e., how well it makes predictions on unseen examples of domains already seen during training. 
Moreover, to the best of our knowledge, no existing works comprehensively study the generalization of the mixture of adapters in the in-domain setting. This literature gap seems counter-intuitive because in-domain evaluation is fundamental and should precede out-of-domain evaluation. 
Moreover, in real-world applications, model owners have incentives to utilize as much as possible available information to improve their models over time.
With the availability of parameter-efficient finetuning methods that are fairly easy to adopt with minimal space and runtime cost, the model owners are then incentivized to quickly fine-tune their models on a few examples collected from a new domain on an additional adapter to optimize the performance (rather than totally relying on out-of-domain prediction capability).
As a result, although in-domain evaluation seems trivial, it is fundamental as one must ensure that the mixture of adapters works well on the tasks they have already been trained on. 
Furthermore, several key questions regarding the resulting mixtures of domain-specific adapters remain unanswered, especially those regarding their generalizability and their adversarial robustness when mixing adapters trained from very different tasks.

\begin{figure}[t]
    \centering
    \includegraphics[width=0.475\textwidth]{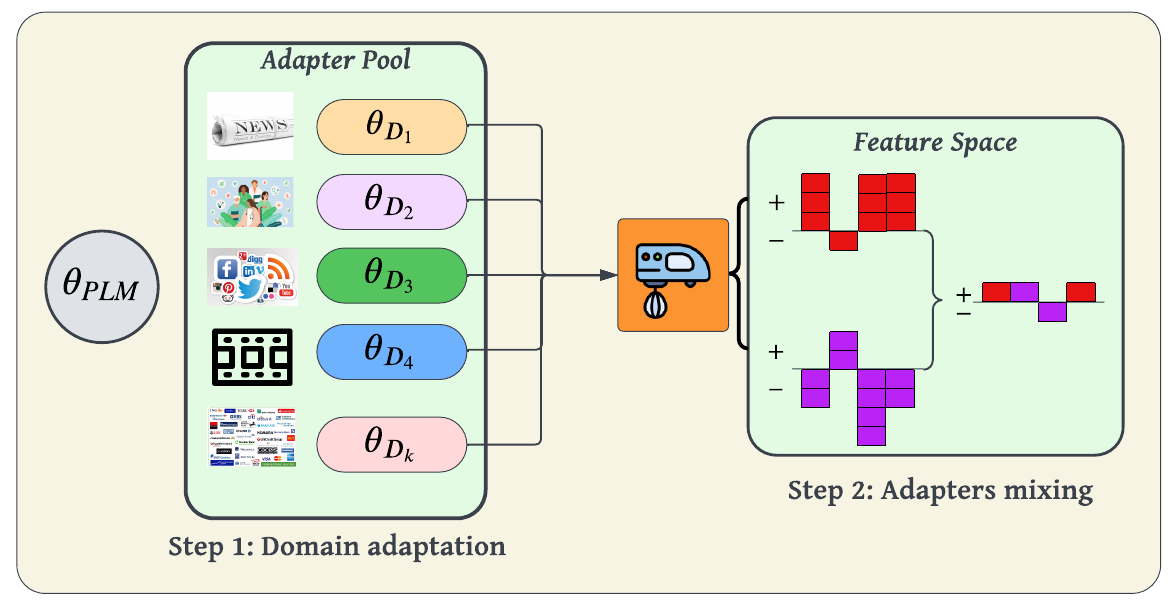}
    \caption{Mixing the adapter weights across various tasks may result in the importance weights of individual tasks nullifying each other, thereby yielding a merged mixture losing important information.
    }
    \label{fig:model_arch}
\vspace{-5pt}
\end{figure}

Therefore, borrowing the pop-culture saying that \textit{``mixed drinks and cocktails aren't actually the same thing''}, in contrast from existing works, we hypothesize that \textit{not all} mixture of expert adapters are created equal and all have superior performance. 
Then, through an array of comprehensive experiments, we attempt to give answers to questions about \textit{when and what to mix} when it comes to domain-specific adapters. 
We found that the weight-space merging mechanism suffers from performance degradation in terms of both generalizability and adversarial robustness even with inputs from domains it already trains on. 
Moreover, we also attempt to explain such performance degradation by revealing a critical negative correlation between \textit{signed directions of adapter weights during mixing} and domain-specific predictive performance (Fig.~\ref{fig:model_arch}). 
Although simple, this intuitive and novel observation also allows us to select \textit{``when and what adapters to mix?''} and design a more effective model pruning as a by-product application. 

Overall, our study does \textit{not} focus on proposing a new mechanism, algorithm, or method.
Instead, we focus on analyzing and bringing understanding of an existing and emerging
paradigm of mixing multiple domain-specific adapters that was previously introduced ~\cite{chronopoulou2023adaptersoup,wang-etal-2022-adamix}. 
Specifically, we focus on in-domain prediction when mixing adapters from different domains as an emerging and potential paradigm for the deployment of PLMs in practice.

Our contributions are summarized as follows.
\begin{enumerate}[leftmargin=\dimexpr\parindent-0.2\labelwidth\relax,noitemsep,topsep=0pt]
    \item This is the first and most comprehensive analysis of in-domain generalizability of a mixture of domain-specific adapters with 3 different adapter methods on 13 diverse classification datasets, 
    \item We provide insights and analysis on when and what adapters to mix to minimize performance degradation via the lens of signed directions of adapters' parameter weights,
    \item We demonstrate the utility of such insights to train mixtures of adapters with 90\% sparsity that improve both generalizability and efficiency.
\end{enumerate}

\section{Related works}\label{sec:related}
\noindent \textbf{Adapter Fine-tuning.} 
The primary method for adapting general-purpose PLMs to downstream tasks is via \emph{full fine-tuning}, which requires adjusting all models' parameters~\cite{peters2018deep,devlin2019bert}. 
However, this results in redundant copies of fine-tuned models for each task, posing a significant memory challenge. 
Thus, various \textit{parameter-efficient fine-tuning methods} have been proposed, including prompt-based~\citep{Li2021Prefix} and adapter-based fine-tuning~\cite{houlsby2019parameter, pfeiffer2020adapterfusion, hu2021lora}. 
Among adapter-based fine-tuning methods, Houlsby~\cite{houlsby2019parameter} introduces two adapter blocks with bottleneck networks in each Transformer block of a PLM. 
Similarly, the Pfeiffer~\cite{pfeiffer2020adapterfusion} adapter differs in architecture, incorporating only one adapter layer in each Transformer block, in contrast to the two layers introduced by Houlsby~\cite{houlsby2019parameter}. 
LoRA~\cite{hu2021lora} takes a distinctive approach by freezing the MLP modules of transformers and representing updates to attention weights with two low-rank matrices to optimize space while effectively retaining model performance. 
In this work, we focus on analyzing adapter-based fine-tuning methods as they are more popular and effective.

\noindent \textbf{Mixture of Expert Adapters.}
Additionally, several approaches~\cite{wang2020k,pfeiffer2020adapterfusion,pfeiffer-etal-2020-mad,wang-etal-2022-adamix} have been proposed to further optimize their adapters for various downstream tasks by maintaining a set of adapters and combine them during inference. 
Particularly, AdaMix~\citet{wang-etal-2022-adamix} fine-tunes so-called Mixture of Experts (MoEs) with adapters on a downstream task and averaging their weights during inference. 
In addition, \citet{li2022branch} explores performance in novel domains through weight averaging on entire language models. 
Similarly, AdapterSoup~\cite{chronopoulou2023adaptersoup} opts for weight-space averaging of adapters trained on different domains. 
Among these methods, weight-space averaging is identified as the most intuitive method for mixing different adapters~\cite{jin2022dataless} and \cite{chronopoulou2023adaptersoup}. 

Nevertheless, none of these works comprehensively evaluates and analyzes the generalizability and adversarial robustness of the resulting mixture of adapters under different mixtures of domain-specific knowledge, which is necessary to answer the question \textit{``when and what to mix?''}. 

\section{Comprehensive In-Domain Evaluation}
To evaluate the in-domain performance of adapter mixtures, we train several adapters with domain-specific knowledge and mix them in different combinatorial combinations. 
Then, we evaluate each combination on different downstream tasks on two aspects: (1) \textit{generalizability} on unseen in-domain examples and (2) \textit{adversarial robustness} under adversarial text attacks.
\begin{figure}[tb!]
    \centering
    \includegraphics[width=0.475\textwidth]{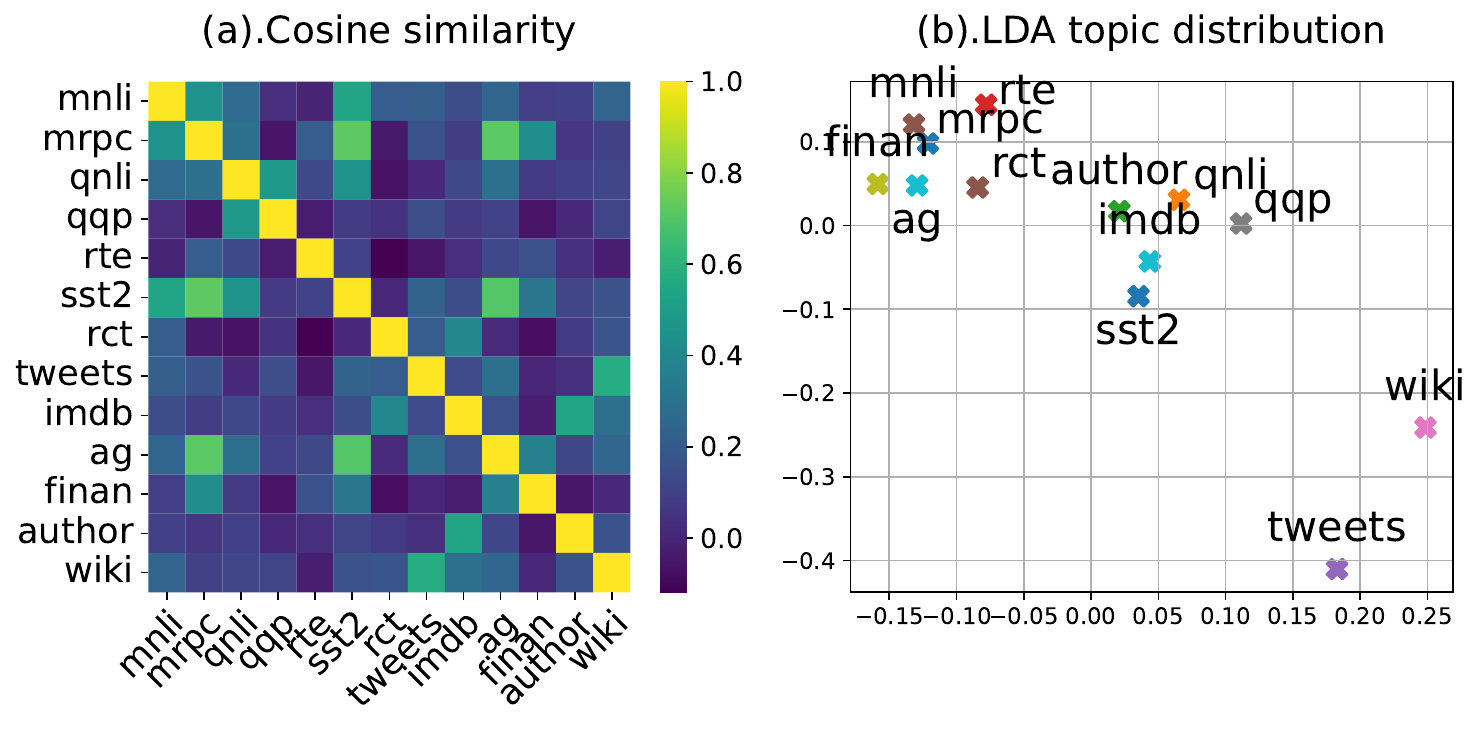}
    \caption{(a) Datasets' semantic similarity via cosine-similarity among centroids of Universal Sentence Encoder (USE)~\cite{cer2018universal} embeddings of 1K randomly sampled documents from each dataset. 
    (b) Topic distributions via Latent Dirichlet Allocation (LDA)~\cite{blei2003latent}.}
    \label{fig:cosine_similarity}
    \vspace{-10pt}
\end{figure}

\subsection{Evaluation Datasets}\label{sec:dataset}
\noindent \textbf{Diverse Domain Knowledge.}
To simulate knowledge diversity, we gather a total of 13 distinct and diverse \textit{domain-specific} datasets of classification tasks for evaluation. 
We refer the readers to Appendix \ref{diverse_knowledge_dataset} for detailed information and their linguistic statistics. 
Fig.~\ref{fig:cosine_similarity} reveals the intricate diversity within our selected datasets, both semantic and topic-wise.
Notably, SST2 and IMDB, both originating from the same movie corpus, exhibit proximity in topic embedding spaces. On the contrary, non-formal datasets such as Wiki and Tweets are distinctly distant from other datasets in this regard. 
We refer the readers to Appendix.~\ref{topic_distribtuion_lda} for a detailed exploration and analysis of the topic distributions among the datasets.

\subsection{Mixing Fine-Tuned Adapters}
\noindent \textbf{Base Models and Individual Adapters.} 
We design our evaluation using two transformer-based models, namely {BERT}~\cite{devlin2018bert} and {RoBERTa}~\cite{liu2019roberta}, with a 3 diverse and well-known adapter methods. 
They are Houlsby~\cite{houlsby2019parameter}, Pfeifer~\cite{pfeiffer2020adapterfusion} and LoRA~\cite{hu2021lora}.
These adapter-based methods introduce variations in the adapter architecture and parameterization (Sec.~\ref{sec:related}), contributing to the comprehensiveness of our analysis. 

\setlength{\belowdisplayskip}{3pt}
\setlength{\abovedisplayskip}{0pt}
\vspace{2pt}
\noindent \textbf{Mixing Adapters.} 
From the pre-trained weights  $\theta_{PLM}$ of either BERT and RoBERTa, we train a suite of 13 domain-specific adapters tailored for diverse domains, denoted as $\theta_{D_1}, \theta_{D_2}, \dots, \theta_{D_k}$. 
Following \cite{chronopoulou2023adaptersoup}, the final inference of the target mixture of domain-specific adapters becomes:
\begin{equation}
    \label{mixing_process}
    f(x, \theta_{PLM} + \frac{1}{k} \sum_{i=1}^{i=k} \theta_{D_i})
\end{equation}

\setlength{\belowdisplayskip}{3pt}
\setlength{\abovedisplayskip}{3pt}

\subsection{Adversarial Text Generation}
Textual adversarial attacks are popular in AI robustness research. 
Given a dataset \( D{=}\{(x_i, y_i)\}_{i}^{N}\), where \( x \) represents the sample and \( y \) denotes the ground truth label, a textual adversarial attack aims to attack a PLMs \( f_\theta \) with a classification loss function \( \mathcal{L} \) by perturbing each sample \( x \) with perturbation noise \( \delta \) given a certain budget \( C \): 
\begin{equation}
    \label{text_adv_attack}
    \text{arg max}_{\delta \in C} \mathcal{L}[f_\theta(x + \delta), y],
\end{equation}

Toward evaluating the robustness of a mixture of adapters, we employ \textit{both black-box and white-box} textual attacks to exercise Eq. \ref{text_adv_attack}. 
We utilize the popular \textit{TextFooler} \cite{jin2020bertrobust} as the \textit{black-box attack}, which aims to replace words with synonyms or contextually similar words to deceive PLMs. 
We utilize the well-known \textit{FGSM} \cite{goodfellow2015} as the \textit{white-box attack}, which can efficiently craft adversarial examples by perturbing embedding of text data in the direction of the sign of the gradient of the loss function to the input, thereby exposing vulnerabilities in model robustness.

\begin{figure*}[t]
    \centering
    \includegraphics[width=\textwidth]{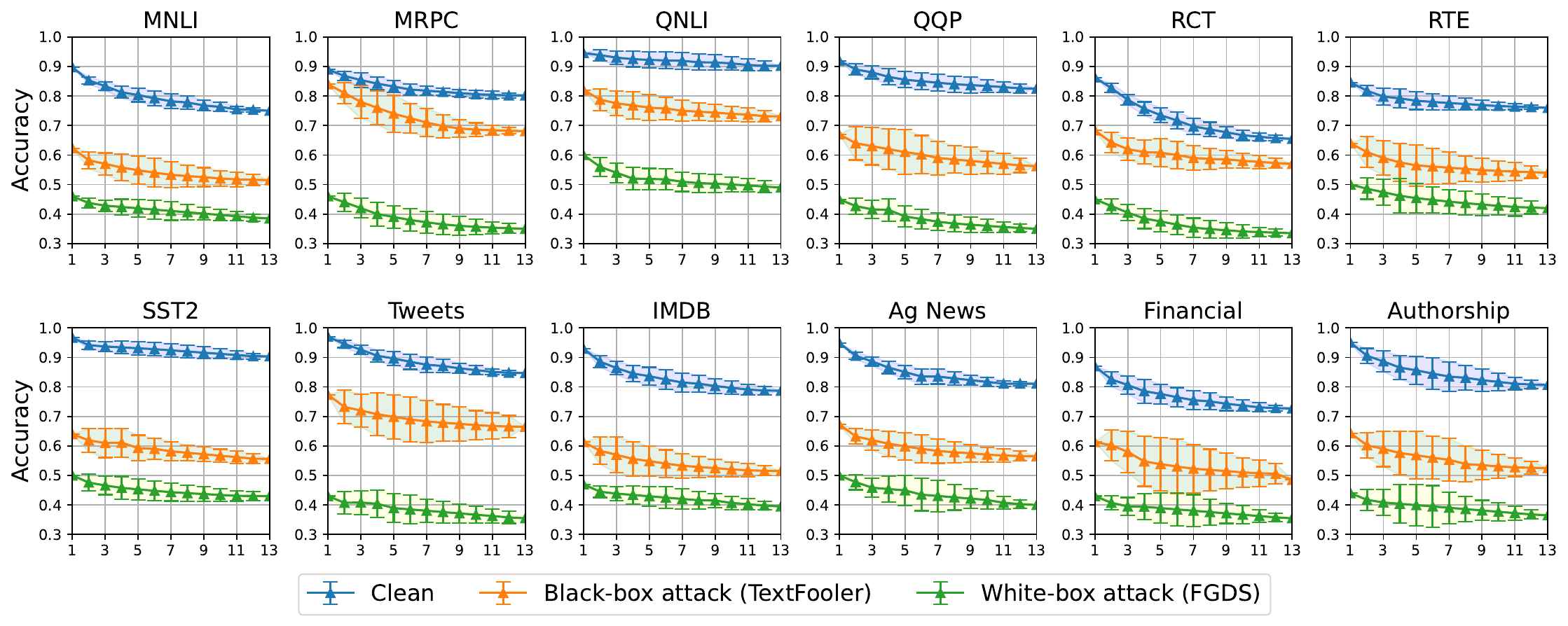}
    \caption{Accuracy of RoBERTa with Pfeiffer~\cite{pfeiffer2020adapterfusion} in each target domain. \textit{X-axis denotes the number of mixed adapters}.}
    \label{roberta_pfeiffer}
\end{figure*}

\renewcommand{\tabcolsep}{2.75pt}
\begin{table*}[t]
\centering
\footnotesize
\begin{tabular}{ccccccccccccccc}
\toprule
     \textbf{Dataset} & \textbf{mnli} & \textbf{mrpc} & \textbf{sst2} & \textbf{rte} & \textbf{qnli}  & \textbf{qqp} & \textbf{rct} & \textbf{ag} & \textbf{authorship} & \textbf{financial} &\textbf{imdb} & \textbf{tweets} & \textbf{wiki} & \textbf{Average}\\
     \midrule
     $\nabla_{clean}$ & \textbf{15.04} & 8.31 & 4.90 &  8.50 & 4.19 & 9.92 & \textbf{21.72} &  \textbf{13.69} & 12.60 & \textbf{15.33} & \textbf{14.26} & \textbf{13.17} & \textbf{13.16} & 11.91 \\ 
     $\nabla_{blackbox}$ & 12.10 & \textbf{16.07} & \textbf{10.43} & \textbf{10.21} & 10.13 & 9.82 & 12.32 &  12.62  & \textbf{13.42} & 13.82 & 13.04 & 11.83 & 10.52 & \textbf{12.03}\\ 
     $\nabla_{whitebox}$ & 10.14 & 11.21 & 8.31  & 9.42 & \textbf{12.14} & \textbf{10.24} & 12.53 & 12.06  & 10.45 & 9.53 & 10.04 & 10.24 & 7.53  & 10.30\\
 \bottomrule
\end{tabular}
\caption{Average \textit{absolute} performance drop (\textit{in percentage \%}) of RoBERTa with Pfeiffer~\cite{pfeiffer2020adapterfusion} when mixed from all domain adapters on clean, black-box, and white-box attacks.} 
\label{performance_drop}  
\vspace{-10pt}
\end{table*}

\subsection{Combinatory Evaluation} 
To evaluate in-domain performance for each target domain, we generate all possible combinations of adapters. 
To illustrate, for the target domain MNLI, we can first evaluate with a mixture of only itself. When combining two adapters, we have the flexibility to choose 1 additional adapter out of the remaining 12, resulting in 12 possible combinations. 
For a set of 3 adapters, including MNLI, we select 2 adapters out of the 12 to generate $C_{12}^2$ combinations. 
This process continues for sets ranging from 4 to 13 adapters, where, in the case of 13 adapters, all adapters are combined.
Thus, we have $13*(1{+}\sum_{i=1}^{12} C_{12}^{i}){=}53,248$ combinations for all domains. 
We report \textit{mean and variance} of in-domain performance for each set of mixtures of $k$ adapters.

Notably, this setup already \textit{includes all the possible mixtures} potentially selected by AdapterSoup~\cite{chronopoulou2023adaptersoup}, which proposes an additional mechanism to select a subset of a fixed number of domain-specific adapters to mix. 

\section{Experiments}
\label{generalizabilty_robustness}
\subsection{Implementation Details}
For the Ag News, Authorship, Financial, IMDB, Tweets, and Wiki-Toxic, we partition the dataset into three segments with an 8:1:1 for train:val:test splits. 
For datasets belonging to the GLUE corpus, we employ their public training and evaluation splits. For the black-box \textit{TextFooler} attack, we set the minimum embedding cosine similarity between a word and its synonyms as 0.8, and the minimum USE similarity is 0.84.
For white-box FGSM~\cite{goodfellow2015} attack, we set the perturbation magnitude to $0.01$.
Following the setup of Houlsby and Pfeiffer~\cite{houlsby2019parameter,pfeiffer2020adapterfusion}, we use all adapters with a dimension of $64$ and $256$ for RoBERTa-large and BERT-base models. 
With LoRA, we use rank $r{=}4$ following \cite{hu2021lora}. 
Detailed training, evaluation dataset, and hyper-parameter configuration for different tasks are presented in Appendix \ref{sec:hyper}.

\subsection{In-Domain Performance Results}
\noindent \textbf{Overview.}  We present the performance of a candidate setting of RoBERTa with Pfeiffer adapter~\cite{pfeiffer2020adapterfusion} with different numbers of additional mixing domains in Fig.~\ref{roberta_pfeiffer}. 
Table \ref{performance_drop} shows how much the predictive performance drops without and with adversarial black-box and white-box attacks when mixing all adapters. 
Overall, the average performance drops over all tasks on the clean test set from the original performance to a mix of all adapters is 11.91\%, and that for black-box and white-box attacks are 12.03\% and 10.30\%, respectively. 
Further results of \textit{generalization and adversarial robustness performance across other models and adapter methods} are documented in Appendix \ref{additional_results_mixing_all}.

\vspace{2pt}
\noindent \textbf{\textit{Finding \#1: As we add more tasks or domains, the predictive performance of every single task decreases}}, reaching its lowest point when we incorporate the maximum of $13$ adapters training on various topic datasets. 
Fig.~\ref{roberta_pfeiffer} shows that mixing domain-specific adapters indeed decreases in-domain performance (reduction of around 10\% in MRPC, QQP, RTE, etc., and nearly 17\% in the Financial domain when mixing 13 adapters). 
The same behaviors were also observed in \cite{jin2023regmean} where they merge the weight of PLMs.
Notably, task accuracies decreased at a slower pace for QNLI and SST2 when evaluating with increasing size of mixtures (Fig.~\ref{roberta_pfeiffer}). 
In contrast, a substantial decrease in accuracy is observed for domains such as RCT, IMDB, Ag News, and Authorship (Fig.~\ref{roberta_pfeiffer}). 
This shows that mixing domain-specific adapters \ul{impair model performance differently} depending on the target domain, or \textit{``what to mix'' in an adapter mixture has a crucial effect on the mixture's performance.} 
To attempt to explain this behavior, we later present and verify a hypothesis that such mixing domain-specific adapters via weight averaging can result in ``forgotten knowledge'' that can happen due to the differences in signs when mixing these adapters (Sec.~\ref{weight_different}). 

\begin{figure}[t]
    \centering
    \includegraphics[width=0.4\textwidth]{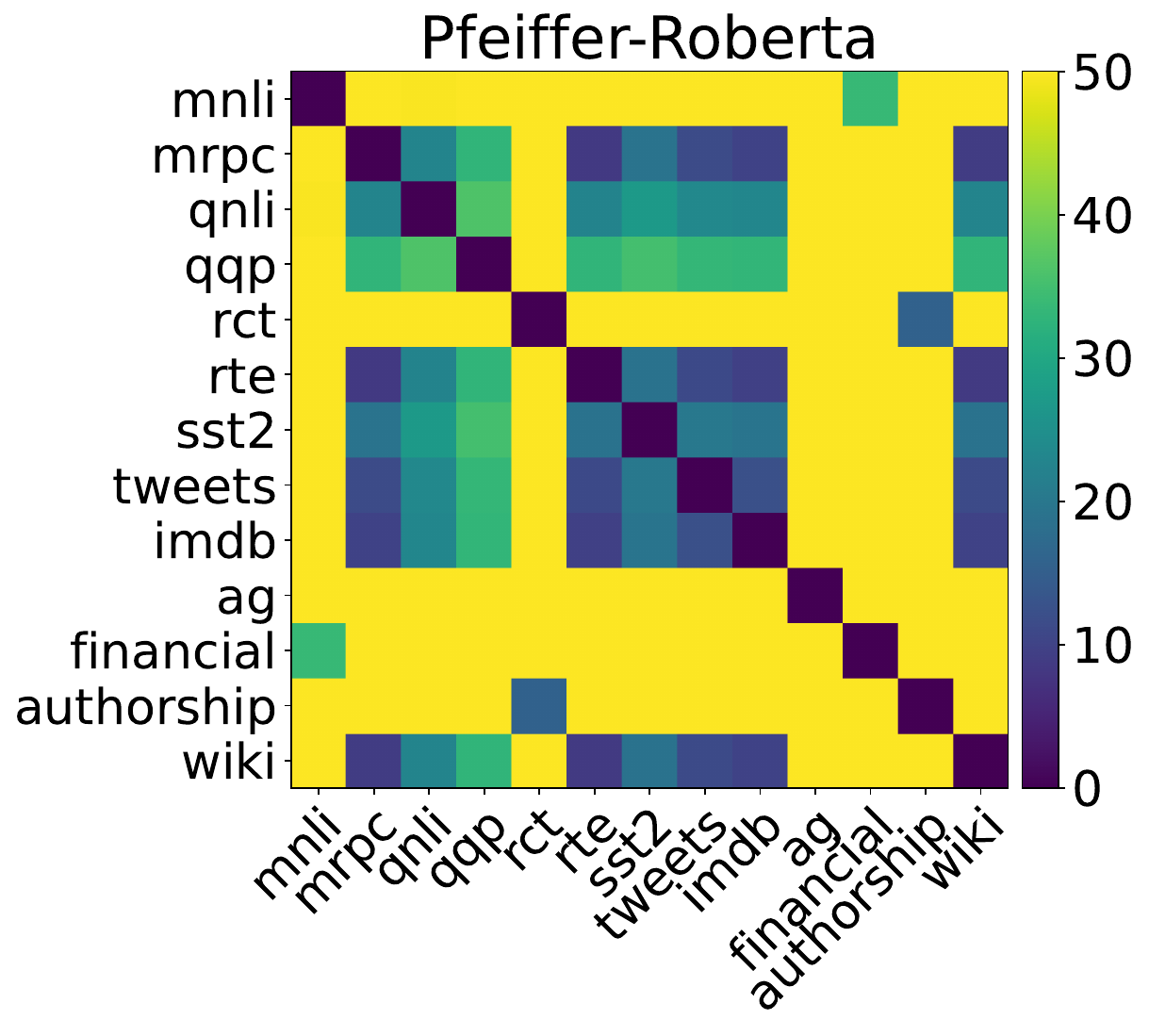}
    \caption{Heatmap visualization of the Fraction of Sign Difference (in \%) of Pfeiffer Adapters~\cite{pfeiffer2020adapterfusion} trained on 13 domain-specific tasks with RoBERTa..}
    \label{weight_different_roberta}
    \vspace{-15pt}
\end{figure}

\vspace{2pt}
\noindent \textbf{\textit{Finding \#2: On average, there are no notable differences of the magnitude in accuracy degradation with and without adversarial attacks (12.03\% dropped in black-box attack versus 11.91\% dropped without attack) (Fig. \ref{roberta_pfeiffer})}.} 
The overall predictive performance was significantly lower under the white-box compared to the black-box attack as expected because the white-box FGSM attack has additional access to the models' parameters. 
Interestingly, on the RCT dataset, the accuracy drop on clean ($21.72\%$) is much larger than on black-box and white-box attacks ($12.32\%$ and $12.53\%$, respectively).

\vspace{2pt}
\noindent \textbf{\textit{Finding \#3: Variance in task accuracy on adversarial attacks, when combining different domain adapters, is observed to be larger than the variance in clean accuracy (Fig. \ref{roberta_pfeiffer}).}} 
Although the magnitude decrease appears similar in Fig.\ref{roberta_pfeiffer}, differences in variance (max, min) are discernible among each combination. 
Specifically, in MRPC, QNLI, QQP, RTE, SST2, Tweets, IMDB, Financial, and Authorship, we can observe that certain mixed models observed slightly better adversarial robustness compared to single adapters (when the number of mixing adapters $k$ is $2$ or $3$). 
Moreover, {the variance of robustness in adversarial scenarios tends to be higher than in clean scenarios because all of the adapters are trained on different tasks and they may exhibit different vulnerabilities to the same attack method}.

\vspace{2pt}
\noindent \textbf{\textit{Finding \#4: Mix up to three adapters to maintain competitive performance.}} Based on the results from Fig.\ref{roberta_pfeiffer} and all findings above, \textit{it is advisable to mix only up to three tasks} to maintain competitive performance (as observed in MRPC, QQP, RTE, Tweet, Financial, and Authorship domains with less than a 3\% accuracy drop in accuracy). 
Especially, in some cases when evaluating QNLI, SST2, mixtures of less than three adapters even achieved better performance in terms of generalizability (up to 1\%) and also in terms of adversarial robustness of up to 3\% compared to the original performance.

\section{Effects of Sign Differences of Adapter Weights during Mixing: A Hypothesis}
\label{weight_different}
\subsection{An Explanation Hypothesis}
The ideal scenario when averaging adapter weights during mixing is to make minimal adjustments to their weights, both in terms of values--i.e., magnitudes, and directions, to sustain as much as possible the knowledge learned. 
Investigating how adapter weights mix regarding both their magnitudes and directions, can be overly complex. 
Thus, we simplify this assessment by focusing on the \textit{sign directions} of the adapter weights in our analysis. Following the mixing process of $k$ individual adapter weight in Eq. \ref{mixing_process} (Sec.~\ref{generalizabilty_robustness}), {{we then hypothesize that mixing \ul{adapter weights of conflicting signs} can result in ``forgotten knowledge'', and lead to performance degradation. 
As illustrated in Fig.~\ref{fig:model_arch}, averaging adapter weights across various tasks may lead to nullifying importance weights for individual tasks if their signs are opposite--i.e., positive v.s. negatives. 
In other words, the \textbf{fraction of sign difference or {FSD} (\%)} or proportion of weight sign difference in adapter weights during mixing correlate with their mixtures' generalizability}}. Alg. \ref{weight_sign_differnce_algorithm} (Appendix \ref{sign_diff_section}) shows the calculation of FSD.

We evaluate our hypothesis with different cases: (i) individual adapters (mixture with $k{=}1$), (ii) dual adapters ($k{=}2$), and generalize to (iii) multiple adapters ($k{>}2$). 
To demonstrate the utility of our hypothesis, we apply it to improve the generalizability of adapter mixtures and also to derive a more effective model pruning in Sec.~\ref{sec:application1}, \ref{application}.

\subsection{Individual Adapters ($k{=}1$)}
\label{individual_adapter}
We calculate the FSD of adapter weights on RoBERTa and normalize it by the total number of weights, denoted as a matrix  $\mathbf{S}_{k\times k}$ where each row is the FSD of a single adapter train on task $k$ to the remaining adapters (Fig.~\ref{weight_different_roberta}). 
We refer the readers to Sec.~\ref{additional_result_bert_diff} in the Appendix for results on BERT. 
Interestingly, a consistent trend in the FSD is observed across various model architectures (BERT, RoBERTa) and adapter methods (e.g., Fig.~\ref{weight_different_roberta} and Fig.~\ref{weight_different_bert} of Appendix~\ref{additional_result_bert_diff}). 
The reason is that adapters act as small MLP layers that integrate task-specific knowledge into pre-trained models \cite{meng2022locating} in different adapter methods. 
This shared functionality contributes to a similar trend in FSD, highlighting the robustness and generalizability of the observed behavior across different adapters' architectural variations.
In addition, Adapters trained on datasets with distinct topic distributions and cosine similarities (Fig.~\ref{fig:cosine_similarity}) exhibit varying weight directions (Fig.~\ref{weight_different_roberta}).  
Especially, MNLI has a similar linguistic distribution with other datasets (i.e. MRPC, RTE, Ag News, etc) (Fig.~\ref{fig:cosine_similarity}), but the adapter trained on MNLI has a significantly larger FSD compared to the remaining domain-specific adapters (Fig.~\ref{weight_different_roberta}).
Thus, \textit{datasets that are similar in linguistic statistics may not necessarily share the same optimization trajectory}.
As a result, methods that are based on the linguistic distribution to choose the closest set of adapters to mix like AdapterSoup \cite{chronopoulou2023adaptersoup} may lead to sub-optimal performance. 

\begin{figure}[t]
    \centering
    \includegraphics[width=0.5\textwidth]{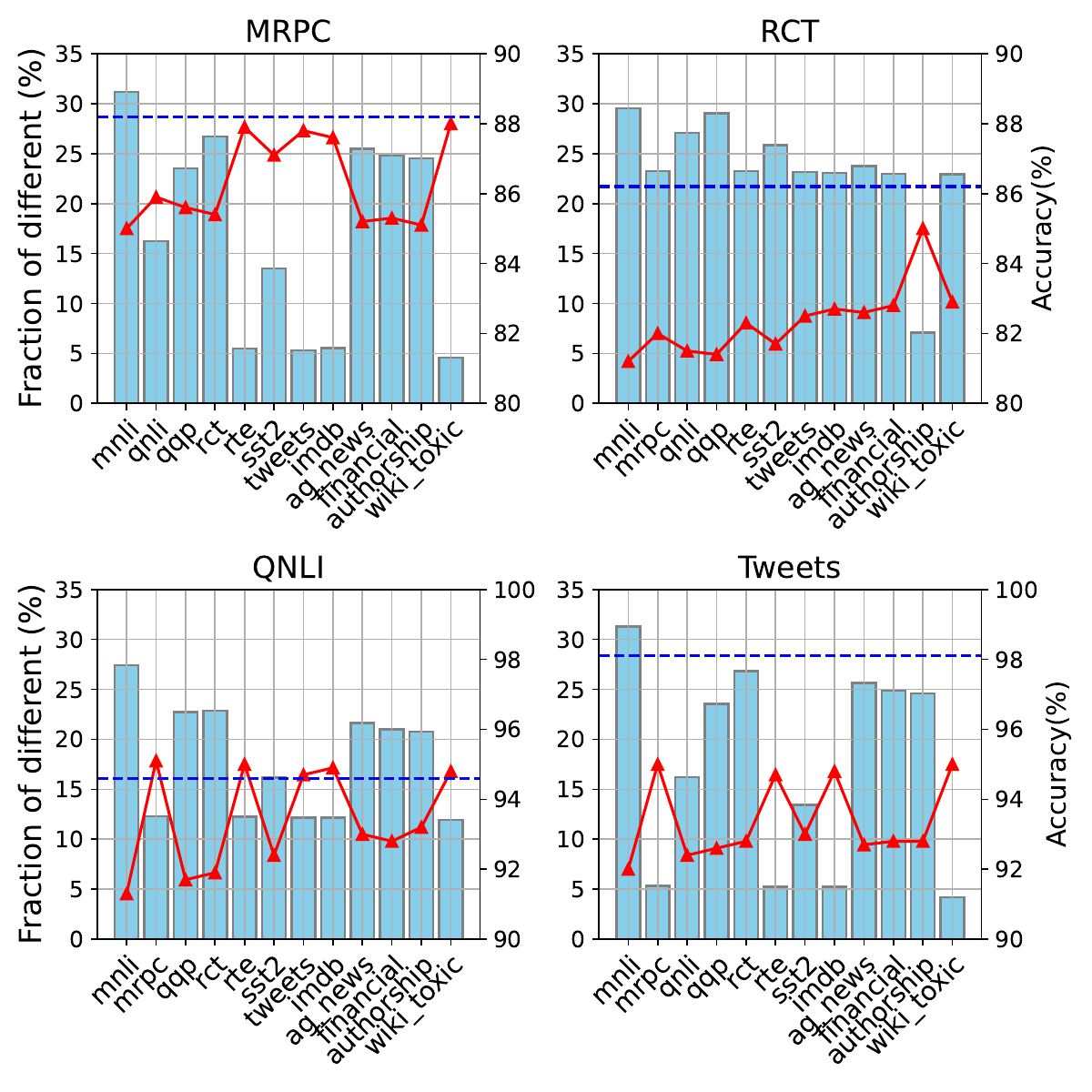}
    \caption{FSD when mixing two ($k{=}2$) adapters. \textcolor{lightskyblue}{Sky-blue} bars show the FSD (left y-axis). \textit{Dashed \textcolor{blue}{blue} lines} denote the accuracy achieved by a standalone adapter. \textit{Solid \textcolor{red}{red} lines} illustrate the variations in accuracy after mixing. Please refer to Fig. \ref{pfeiffer_roberta_two_adapters_diff_full} in Appendix \ref{two_adapter_diff_full} for results in other tasks.}
    \label{pfeiffer_roberta_two_adapters_diff}
    \vspace{-10pt}
\end{figure}

\subsection{Dual Adapters ($k{=}2$)} 
\label{dual_adapters}
Fig.~\ref{pfeiffer_roberta_two_adapters_diff} shows the FSD for each adapter when mixing two domain-specific adapters. 
This investigation is conducted within the context of the Pfeiffer Adapter~\cite{pfeiffer2020adapterfusion} using a pre-trained RoBERTa model. 
Overall, there is a strong negative correlation between the FSD and the generalizability--i.e., the lower the sky-blue bar (or the smaller fraction of weight sign conflicts), the higher the performance of the mixture (Fig.~\ref{pfeiffer_roberta_two_adapters_diff}).
Notably, \textit{tasks with substantial difference in the weight signs witness a pronounced performance decrease}. 
Specifically, RCT exhibits significant performance drops due to substantial differences in adapter weight direction. 
Conversely, tasks such as MRPC, and QNLI demonstrate either marginal improvement or no change in performance when mixed with other adapters. 
This is well correlated to the marginal FSD of the mixed adapters, ranging from only 5\% to 10\% compared to the original weight.

\begin{figure}[t]
    \centering
    \hspace*{-7pt}
    \includegraphics[width=0.5\textwidth]{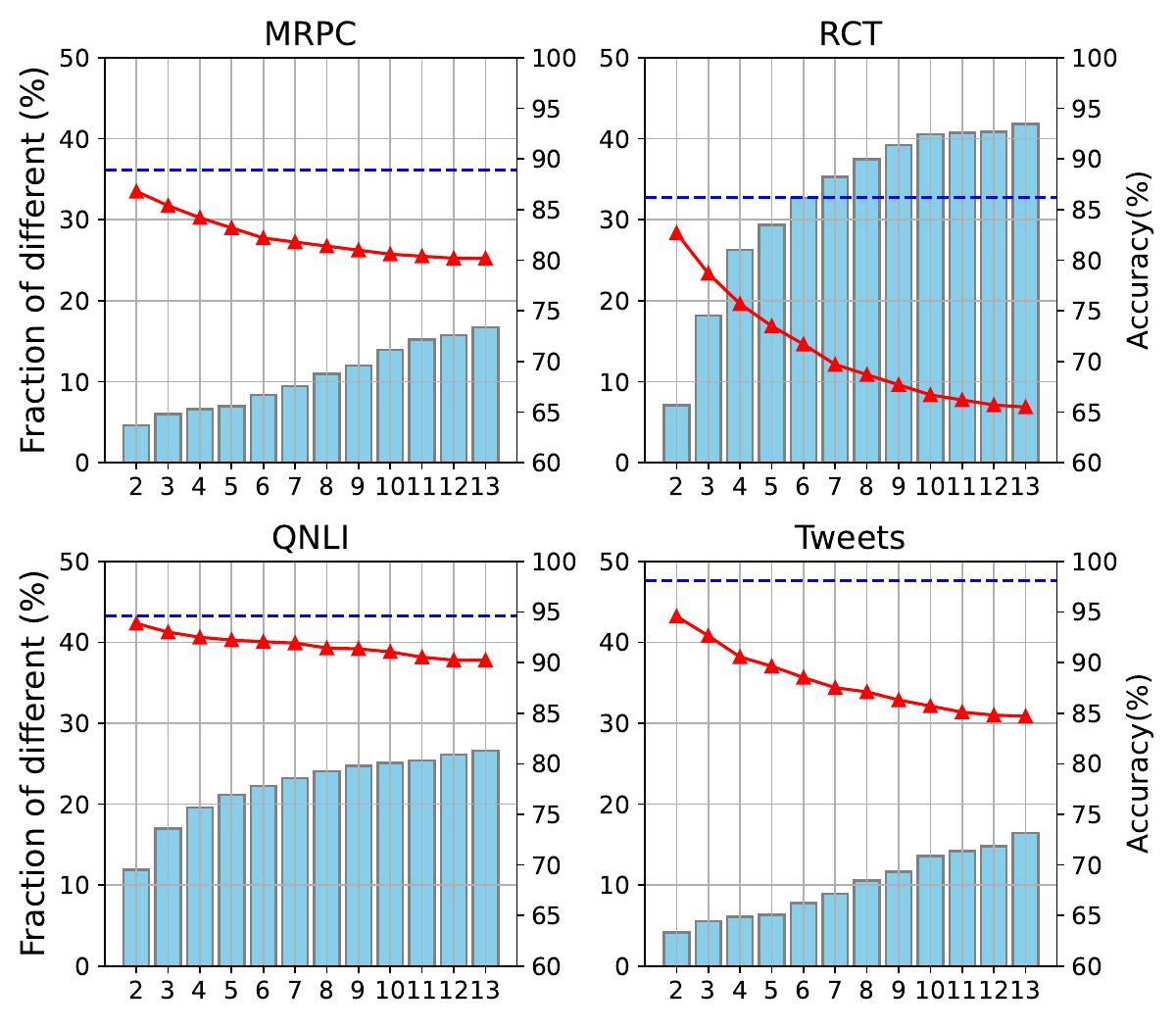}
    \caption{Fraction of weights changing direction during the mixing of multiple adapters, ranging from 2 to 13.
    The chart description is similar to Fig.~\ref{pfeiffer_roberta_two_adapters_diff}.
    We refer to Fig. \ref{pfeiffer_roberta_multi_adapters_diff_full} in Appendix \ref{two_adapter_diff_full} for detailed results in other tasks.}
    \label{pfeiffer_roberta_multi_adapters_diff}
    \vspace{-10pt}
\end{figure}

\subsection{Multiple Adapters ($k{>}2$)} \label{multiple_adapters}
{Similar to the dual-adapter setting}, there is \textit{still a strong negative correlation between FSD and the generalizability}, and increasing the number of mixed adapters amplified the sign disparity (Fig.~\ref{pfeiffer_roberta_multi_adapters_diff}). 
For example, adapters trained on RCT and Tweet exhibit large FSD compared to other adapters and hence observed a significant decrease in generalizability (Fig.~\ref{pfeiffer_roberta_multi_adapters_diff}). 
In some cases, mixing only a few adapters (e.g., 2--4) could still maintain competitive performance as in QNLI, SST2 domains (Fig. \ref{pfeiffer_roberta_multi_adapters_diff}, \ref{pfeiffer_roberta_multi_adapters_diff_full}).
This correlates with the relatively marginal differences in adapter weights of QNLI and SST compared to adapters trained on other domains (Fig.~\ref{weight_different_roberta}), as their mixtures do not lead to significant cancellations of existing parameters and hence preserve learned knowledge.

\begin{figure}[tb]
    \centering
    \hspace*{-7pt}
    \includegraphics[width=0.5\textwidth]{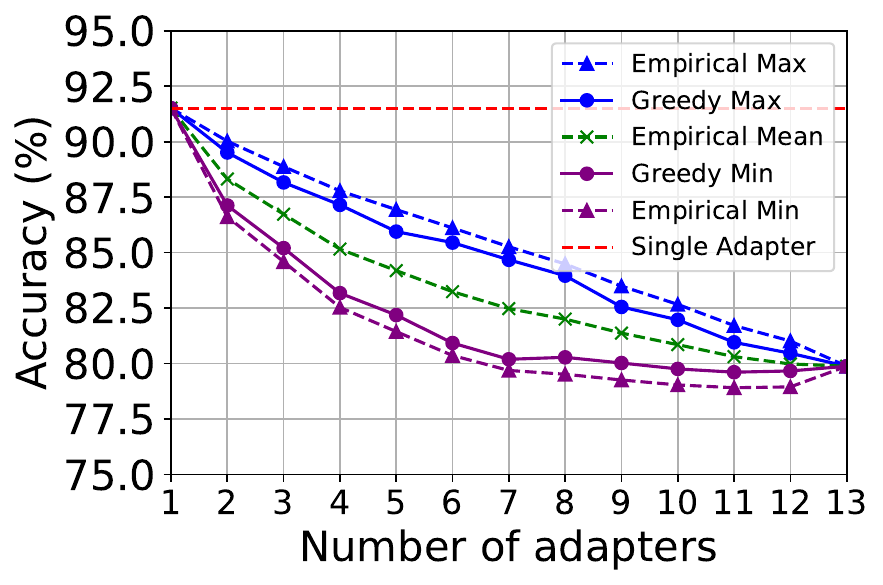} 
    \caption{Average model accuracy of 13 domains under different numbers of mixed adapters ($l$) using the guidance of FSD.}
    \label{greedy_adapter_mixing}
    \vspace{-10pt}
\end{figure}

\subsection{Discussion}
\noindent \textbf{Which adapters we should mix?} From Sec.~\ref{weight_different}, we find that it is not advisable to merge adapters that are significantly different from each other in weight signs. 
On the other hand, mixing the weight of these adapters with other adapters which have a small FSD achieves competitive performance (within 3\% drop in accuracy) \ul{only} when the number of mixing adapters is small (QNLI, MRPC in Fig.~\ref{pfeiffer_roberta_multi_adapters_diff}) or can achieve better performance, although rarely, compared to training the original adapter as seen in QNLI domain (Fig.~\ref{pfeiffer_roberta_two_adapters_diff}).
Therefore, when deploying PLMs, it is prudent to only select a group of tasks with a small FSD to minimize the performance drop in the final mixed model.
Our observation is crucial in deploying these models in edge devices where only the adapters are saved on edge, which often has a specific memory capacity limit.

\vspace{2pt}
\noindent \textbf{Experiment comparison with AdapterSoup.} AdapterSoup \cite{chronopoulou2023adaptersoup} dynamically selects a set of $l$ adapters during inference. When $l{=}k$, then our experiment setting is the same as the AdapterSoup setting when we use all adapters at once. 
When $l{=}1$, it is the original performance of a single adapter, assuming that AdapterSoup is perfect at picking the same domain that is already trained on, and this result is already included in our paper (mixture of only $1$ adapter, Fig.~\ref{roberta_pfeiffer}). 
When $1{<}l{<}k$, we do not have the results of the specific combination that AdapterSoup would select. 
However, we reported the maximum performance across all combinatorial combinations among 13 domain-specific adapters or each value in Fig.~\ref{roberta_pfeiffer}. 
For example, when $l{=}3$, we only observe a possible comparative performance (within less than 3\% drop) in QNLI, MRPC, and SST2 domains (Fig. \ref{pfeiffer_roberta_multi_adapters_diff_full}). 
We also emphasize that it is one thing to select the best adapters to mix during inference, it is much harder to choose $l$ or how many of them.

\begin{algorithm}[tb]
\footnotesize
\caption{Greedy Adapter Mixing}
\label{greedysoupalgorithm}
\textit{\ul{\textbf{Input:}}} $k$ domain-specific adapters, a matrix  $\mathbf{S}_{k\times k}$ of FSD, $l$ number of adapters to fusion.\\
\textit{\ul{\textbf{Output:}}} Average($\mathsf{candidates}$) \\
\begin{algorithmic}[1]
\State{$\mathsf{candidates} \gets \{ \}$}
\State{Compute the average of FSD \\ $avg_S = mean(S, axis = 1)$}
\State{Select the top $l$ from set of $k$ adapters according to \\ smallest average FSD}
\State ${\mathsf{candidates} \gets \mathsf{top}_l}$
\State {\bfseries return} $\operatorname{average}(\operatorname{\mathsf{candidates}})$
\end{algorithmic}
\end{algorithm}

\section{Greedy Adapter Mixing with FSD} \label{sec:application1}
Our observations from Sec.~\ref{weight_different} reveal that two adapters with minimal disparity in FSD can yield competitive performance when combined. 
Therefore, to demonstrate the utility of FSD, in this section, we design a mixing strategy, so-called \textit{Greedy Adapter Mixing} (Alg. \ref{greedysoupalgorithm}), that utilizes FSD to decide which domain-specific adapters to mix by minimizing the overall FSD in a greedy manner to get a final mixed model with competitive performance. 
This algorithm is also based on the hypothesis that mixing adapters with minimal FSD can yield mixture models of better generalizability. 
We proceed by evaluating model performance across various adapter combinations. 

Fig.~\ref{greedy_adapter_mixing} shows the generalizability performance of a mixture of $l\in[2,13]$ domain-specific adapters. 
Overall, Greedy Adapter Mixing resulted in very competitive performance compared with empirical upper-bound accuracy. 
In contrast, using the same algorithm \ul{but} maximizing FSD resulted in performance close to the empirical lower-bound accuracy (Fig.~\ref{greedy_adapter_mixing}).
However, in both two cases, mixing adapters with FSD cannot achieve the empirical upper-bound and lower-bound performance.
Thus, \textit{greedily mixing adapters to minimize FSD does not totally prevent knowledge loss in the adapter mixtures}. 
Nevertheless, FSD is still useful as a guidance measure to effectively mix domain-specific adapters.

\section{Towards Effective Model Pruning}
\label{application}
To further demonstrate the utility of our FSD analysis in Sec.~\ref{weight_different} and Sec.~\ref{sec:application1}, in this section, we leverage FSD information to reduce knowledge loss through the development of a pruning algorithm guided by FSD insights.
Specifically, Fig.~\ref{pfeiffer_roberta_two_adapters_diff} shows that predictive performance experiences a significant drop when integrating adapters with pronounced disparities in weight signs--i.e., positive v.s. negative signs. 
Moreover, neural network pruning indicates that only a limited number of weights significantly contribute to task performance, suggesting redundancy within the weights that can be pruned without compromising the original task performance~\cite{han2015deep,frankle2020pruning, lazarevich2021post}. 
Thus, to mitigate the impact of weight sign differences in adapter mixtures, we propose mixing only the \textit{sparse versions} of the adapters' weights. 

Different from Alg. \ref{greedysoupalgorithm}, this strategy \textit{indirectly reduces the fraction of weight sign conflicts}. 
Intuitively, by minimizing the FSD, the mixing process becomes more resilient to the inadvertent elimination of important weights by less significant or redundant weights. 
This phenomenon is visually depicted in Step 2 of Fig.~\ref{fig:model_arch}, where only significant weights in the two adapters need to be preserved, and small or unimportant weights of opposing signs can be eliminated.

\subsection{FSD-based Magnitude Pruning}
\label{magnitude_pruning_alg}

\begin{algorithm}[tb!]
\footnotesize
\caption{FSD-based Pruning}
\label{pruning}
\textit{\ul{\textbf{Input:}}} adapter parameters $w$, sparse ratio $s$. \\
\textit{\ul{\textbf{Output:}}} pruned adapter $\tilde{w}$. \\
\begin{algorithmic}[1]
\State{ $w \leftarrow \operatorname{Trained}(w)$}
\State{ Compute important score $z = |w|$}
\State{ Compute the $s$-th percentile of $z$ as $z_s$}
\State{ $m \leftarrow \mathbbm{1} \left[z - {z_s} \geq 0\right]$}
\State{ $\tilde{w} \leftarrow m \odot w$}
\end{algorithmic}
\end{algorithm}

\vspace{2pt}
\noindent \textbf{Post-training Pruning.}
Sparse Adapter~\cite{he-etal-2022-sparseadapter} employs pruning across every layer of adapters, being able to achieve comparable or even superior predictive performance than standard adapters, even when the sparse ratio reaches up to 80\%. By adopting a similar process of adapters pruning after training, \textit{with the guidance of our FSD analysis in Sec.~\ref{weight_different}}, we can \textit{eliminate redundant parameters at an early stage, circumventing the need for a time-consuming iterative pruning process}, as discussed in prior works such as \citet{han2015deep, lazarevich2021post}. \textit{The detailed pruning algorithm is presented in Alg.~\ref{pruning}}.

\begin{figure}[tb]
    \centering
    \hspace*{-7pt}
    \includegraphics[width=0.5\textwidth]{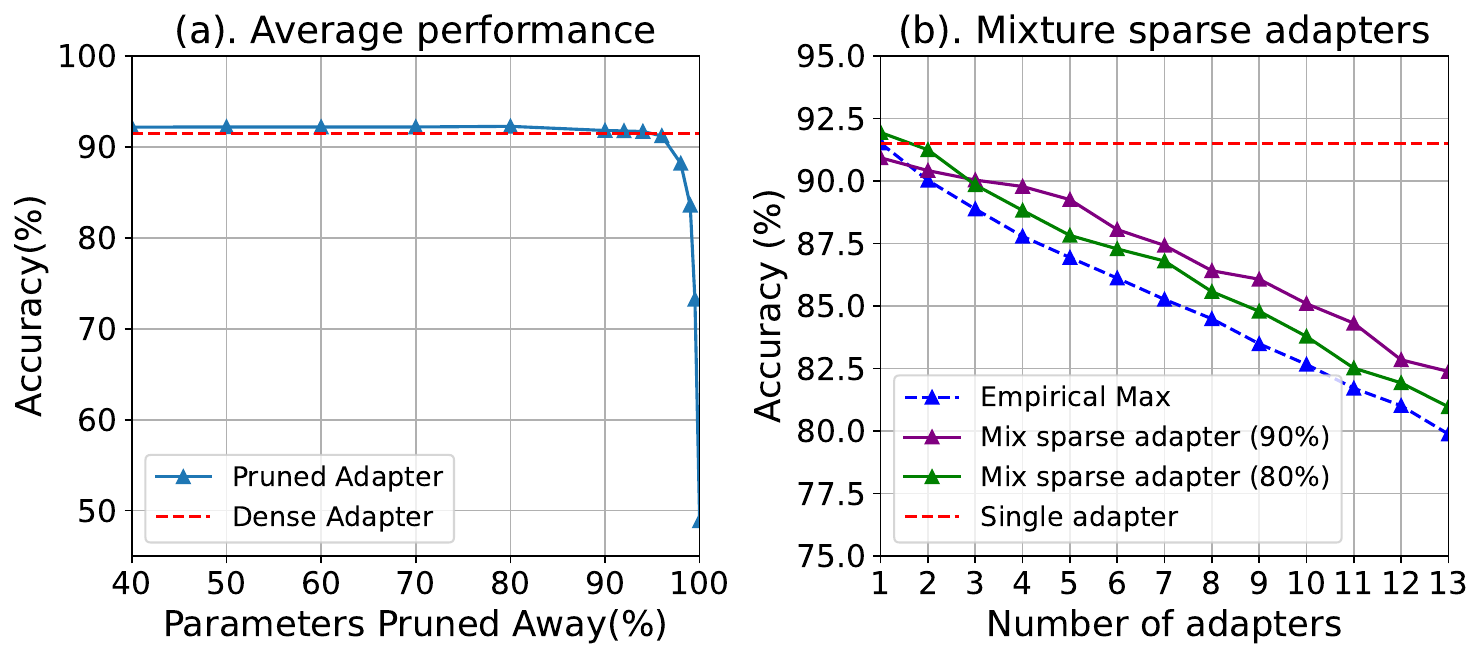}
    \caption{(a) Average RoBERTa performance with a single sparse adapter across 13 domains with increasing sparsity. 
    (b) Model accuracy when increasing the \# of sparse adapters being mixed. \textit{\textcolor{red}{Dashed red}} lines represent model generalization when mixing domain-specific adapters. \textit{\textcolor{blue}{Dashed blue}} lines depics the maximum performance when mixing $k$ domain-specific dense adapters. \textit{\textcolor{green}{Solid green}} and \textit{\textcolor{purple}{solid purple}} lines represent model performance when mixing $k$ domain-specific adapters with 80\% and 90\% sparsity, respectively.}
    \label{mixing_sparse_adapter}
\end{figure}

\vspace{2pt}
\noindent \textbf{Generalizability of Pruned Adapters.}
Fig.~\ref{mixing_sparse_adapter}a shows RoBERTa's performance, where we systematically prune the weight of the Pfeiffer adapter 40\%--100\% of sparsity. For a single task at the sparsity level $d\%$, we retain only the largest--i.e., the top-$d\%$, influential parameters of the corresponding adapter, and report the average in-domain performance across 13 domains. 
Remarkably, pruning up to 90\% parameters of adapter weight does not lead to performance degradation. 
This observation suggests redundancy in adapters' parameters may \textit{contribute to the increase in the fraction of weight direction conflicts}--i.e., high FSD when merging them (Fig.~\ref{pfeiffer_roberta_multi_adapters_diff}).

\vspace{2pt}
\noindent \textbf{Generalizability of Mixed Pruned Adapters.}
Motivated by pruned adapter can still maintain original performance up to 90\% of sparsity (Fig. \ref{mixing_sparse_adapter}a), we hypothesize that mixing sparse adapters may indirectly reduce the fraction of weight sign conflict, therefore, leading to competitive performance with the original adapters.
Given a set of domain-specific adapters, based on the FSD, we choose the top $k$ layers with the highest fraction of weight sign difference and prune each layer up to 90\% of sparsity. Then we mix these adapters by weight averaging. Details of mixing domain-specific adapters with weight sign conflict information is shown in Alg.~\ref{mixing_pruned_adapter_with_sign_conflict_information} (Appendix \ref{mixing_sparse_with_sign_diff}). 
Fig.~\ref{mixing_sparse_adapter}b shows that mixing adapters with 80\% or 90\% sparsity consistently achieved better performance than the upper-bound empirical accuracy achieved when mixing their dense versions.

\section{Conclusion}
This work provides a comprehensive empirical in-domain evaluation of the emerging mechanism of mixing domain-specific adapters. We also provide insights into the inner workings of the mixture of domain-specific adapters by analyzing their weight signs, yielding critical observations on the negative correlation between the fraction of sign difference among adapters and their mixtures' generalizability. By examining the signed directions of adapter weights, we also offer the readers valuable advice on the optimal selection of adapters to mix to achieve competitive performance. Such examination also helps enhance our understanding of the interconnected role of weight sign difference in the context of sparse neural networks.

\section*{Limitation}
Primarily, our exploration focused solely on one classic pruning method, namely Magnitude Pruning~\cite{sanh2020movement} while there are existing more advanced pruning techniques such as SynFlow~\cite{tanaka2020pruning}, GraSP~\cite{wang2020picking} that are also applicable for condensing neural network architectures. 
Consequently, future works include investigating the applicability of our findings to these alternative pruning approaches. 
Furthermore, our examination was confined to the natural language understanding tasks. 
A valuable avenue for future research would involve extending our analysis to encompass the emerging text generation tasks, particularly within the context of the current transformer-based language model, including but not limited to the machine translation tasks utilizing complex GPT-family models.

\bibliography{anthology,custom}

\begin{thebibliography}{39}
\expandafter\ifx\csname natexlab\endcsname\relax\def\natexlab#1{#1}\fi

\bibitem[{Altakrori et~al.(2021)Altakrori, Cheung, and Fung}]{altakrori2021topic}
Malik Altakrori, Jackie Chi~Kit Cheung, and Benjamin~CM Fung. 2021.
\newblock The topic confusion task: A novel evaluation scenario for authorship attribution.
\newblock In \emph{EMNLP}.

\bibitem[{Bentivogli et~al.(2009)Bentivogli, Clark, Dagan, and Giampiccolo}]{bentivogli2009fifth}
Luisa Bentivogli, Peter Clark, Ido Dagan, and Danilo Giampiccolo. 2009.
\newblock The fifth pascal recognizing textual entailment challenge.
\newblock In \emph{TAC}.

\bibitem[{Blei et~al.(2003)Blei, Ng, and Jordan}]{blei2003latent}
David~M Blei, Andrew~Y Ng, and Michael~I Jordan. 2003.
\newblock Latent dirichlet allocation.
\newblock In \emph{Journal of machine Learning research}.

\bibitem[{Cer et~al.(2018)Cer, Yang, Kong, Hua, Limtiaco, John, Constant, Guajardo-Cespedes, Yuan, Tar et~al.}]{cer2018universal}
Daniel Cer, Yinfei Yang, Sheng-yi Kong, Nan Hua, Nicole Limtiaco, Rhomni~St John, Noah Constant, Mario Guajardo-Cespedes, Steve Yuan, Chris Tar, et~al. 2018.
\newblock Universal sentence encoder for english.
\newblock In \emph{EMNLP}.

\bibitem[{Chronopoulou et~al.(2023)Chronopoulou, Peters, Fraser, and Dodge}]{chronopoulou2023adaptersoup}
Alexandra Chronopoulou, Matthew~E Peters, Alexander Fraser, and Jesse Dodge. 2023.
\newblock Adaptersoup: Weight averaging to improve generalization of pre-trained language models.
\newblock In \emph{EACL}.

\bibitem[{Dernoncourt and Lee(2017)}]{dernoncourt-lee-2017-pubmed}
Franck Dernoncourt and Ji~Young Lee. 2017.
\newblock {P}ub{M}ed 200k {RCT}: a dataset for sequential sentence classification in medical abstracts.
\newblock In \emph{Proceedings of the Eighth International Joint Conference on Natural Language Processing}.

\bibitem[{Devlin et~al.(2019{\natexlab{a}})Devlin, Chang, Lee, and Toutanova}]{devlin2019bert}
Jacob Devlin, Ming-Wei Chang, Kenton Lee, and Kristina Toutanova. 2019{\natexlab{a}}.
\newblock Bert: Pre-training of deep bidirectional transformers for language understanding.
\newblock In \emph{NAACL}.

\bibitem[{Devlin et~al.(2019{\natexlab{b}})Devlin, Chang, Lee, and Toutanova}]{devlin2018bert}
Jacob Devlin, Ming-Wei Chang, Kenton Lee, and Kristina Toutanova. 2019{\natexlab{b}}.
\newblock Bert: Pre-training of deep bidirectional transformers for language understanding.
\newblock In \emph{NAACL}.

\bibitem[{Diao et~al.(2023)Diao, Xu, Xu, Wang, and Zhang}]{diao2023mixture}
Shizhe Diao, Tianyang Xu, Ruijia Xu, Jiawei Wang, and Tong Zhang. 2023.
\newblock Mixture-of-domain-adapters: Decoupling and injecting domain knowledge to pre-trained language models memories.
\newblock In \emph{ACL}.

\bibitem[{Dolan and Brockett(2005)}]{dolan2005automatically}
Bill Dolan and Chris Brockett. 2005.
\newblock {Automatically constructing a corpus of sentential paraphrases}.
\newblock In \emph{Third International Workshop on Paraphrasing}.

\bibitem[{Frankle et~al.(2021)Frankle, Dziugaite, Roy, and Carbin}]{frankle2020pruning}
Jonathan Frankle, Gintare~Karolina Dziugaite, Daniel~M Roy, and Michael Carbin. 2021.
\newblock Pruning neural networks at initialization: Why are we missing the mark?
\newblock In \emph{ICLR}.

\bibitem[{Goodfellow et~al.(2015)Goodfellow, Shlens, and Szegedy}]{goodfellow2015}
Ian~J. Goodfellow, Jonathon Shlens, and Christian Szegedy. 2015.
\newblock Explaining and harnessing adversarial examples.
\newblock In \emph{ICLR}.

\bibitem[{Han et~al.(2016)Han, Mao, and Dally}]{han2015deep}
Song Han, Huizi Mao, and William~J Dally. 2016.
\newblock Deep compression: Compressing deep neural networks with pruning, trained quantization and huffman coding.
\newblock In \emph{ICLR}.

\bibitem[{He et~al.(2022)He, Ding, Dong, Zhang, and Tao}]{he-etal-2022-sparseadapter}
Shwai He, Liang Ding, Daize Dong, Jeremy Zhang, and Dacheng Tao. 2022.
\newblock {S}parse{A}dapter: An easy approach for improving the parameter-efficiency of adapters.
\newblock In \emph{EMNLP}.

\bibitem[{Houlsby et~al.(2019)Houlsby, Giurgiu, Jastrzebski, Morrone, De~Laroussilhe, Gesmundo, Attariyan, and Gelly}]{houlsby2019parameter}
Neil Houlsby, Andrei Giurgiu, Stanislaw Jastrzebski, Bruna Morrone, Quentin De~Laroussilhe, Andrea Gesmundo, Mona Attariyan, and Sylvain Gelly. 2019.
\newblock Parameter-efficient transfer learning for nlp.
\newblock In \emph{ICLR}.

\bibitem[{Hu et~al.(2022)Hu, Shen, Wallis, Allen-Zhu, Li, Wang, Wang, and Chen}]{hu2021lora}
Edward~J Hu, Yelong Shen, Phillip Wallis, Zeyuan Allen-Zhu, Yuanzhi Li, Shean Wang, Lu~Wang, and Weizhu Chen. 2022.
\newblock Lora: Low-rank adaptation of large language models.
\newblock In \emph{ICLR}.

\bibitem[{Iyer et~al.(2017)Iyer, Dandekar, Csernai et~al.}]{iyer2017first}
Shankar Iyer, Nikhil Dandekar, Korn{\'e}l Csernai, et~al. 2017.
\newblock First quora dataset release: Question pairs.
\newblock In \emph{data. quora. com}.

\bibitem[{Jin et~al.(2020)Jin, Jin, Zhou, and Szolovits}]{jin2020bertrobust}
Di~Jin, Zhijing Jin, Joey~Tianyi Zhou, and Peter Szolovits. 2020.
\newblock Is {BERT} really robust? {A} strong baseline for natural language attack on text classification and entailment.
\newblock In \emph{AAAI}.

\bibitem[{Jin et~al.(2023{\natexlab{a}})Jin, Ren, Preotiuc-Pietro, and Cheng}]{jin2022dataless}
Xisen Jin, Xiang Ren, Daniel Preotiuc-Pietro, and Pengxiang Cheng. 2023{\natexlab{a}}.
\newblock Dataless knowledge fusion by merging weights of language models.
\newblock In \emph{ICLR}.

\bibitem[{Jin et~al.(2023{\natexlab{b}})Jin, Ren, Preotiuc-Pietro, and Cheng}]{jin2023regmean}
Xisen Jin, Xiang Ren, Daniel Preotiuc-Pietro, and Pengxiang Cheng. 2023{\natexlab{b}}.
\newblock \href {https://openreview.net/forum?id=FCnohuR6AnM} {Dataless knowledge fusion by merging weights of language models}.
\newblock In \emph{ICLR}.

\bibitem[{Lazarevich et~al.(2021)Lazarevich, Kozlov, and Malinin}]{lazarevich2021post}
Ivan Lazarevich, Alexander Kozlov, and Nikita Malinin. 2021.
\newblock Post-training deep neural network pruning via layer-wise calibration.
\newblock In \emph{ICCV}.

\bibitem[{Li et~al.(2022)Li, Gururangan, Dettmers, Lewis, Althoff, Smith, and Zettlemoyer}]{li2022branch}
Margaret Li, Suchin Gururangan, Tim Dettmers, Mike Lewis, Tim Althoff, Noah~A Smith, and Luke Zettlemoyer. 2022.
\newblock Branch-train-merge: Embarrassingly parallel training of expert language models.
\newblock In \emph{arXiv}.

\bibitem[{Li and Liang(2021)}]{Li2021Prefix}
Xiang~Lisa Li and Percy Liang. 2021.
\newblock Prefix-tuning: Optimizing continuous prompts for generation.
\newblock In \emph{ACL}.

\bibitem[{Liu et~al.(2019)Liu, Ott, Goyal, Du, Joshi, Chen, Levy, Lewis, Zettlemoyer, and Stoyanov}]{liu2019roberta}
Yinhan Liu, Myle Ott, Naman Goyal, Jingfei Du, Mandar Joshi, Danqi Chen, Omer Levy, Mike Lewis, Luke Zettlemoyer, and Veselin Stoyanov. 2019.
\newblock Roberta: A robustly optimized bert pretraining approach.
\newblock In \emph{arXiv}.

\bibitem[{Malo et~al.(2014)Malo, Sinha, Korhonen, Wallenius, and Takala}]{malo2014good}
Pekka Malo, Ankur Sinha, Pekka Korhonen, Jyrki Wallenius, and Pyry Takala. 2014.
\newblock Good debt or bad debt: Detecting semantic orientations in economic texts.
\newblock In \emph{Journal of the Association for Information Science and Technology}.

\bibitem[{Matena and Raffel(2022)}]{fisheravg}
Michael Matena and Colin Raffel. 2022.
\newblock \href {https://arxiv.org/abs/2111.09832} {Merging models with fisher-weighted averaging}.
\newblock In \emph{NeurIPS}.

\bibitem[{Meng et~al.(2022)Meng, Bau, Andonian, and Belinkov}]{meng2022locating}
Kevin Meng, David Bau, Alex Andonian, and Yonatan Belinkov. 2022.
\newblock Locating and editing factual associations in gpt.
\newblock In \emph{NeurIPS}.

\bibitem[{Peters et~al.(2018)Peters, Neumann, Iyyer, Gardner, Clark, Lee, and Zettlemoyer}]{peters2018deep}
Matthew~E Peters, Mark Neumann, Mohit Iyyer, Matt Gardner, Christopher Clark, Kenton Lee, and Luke Zettlemoyer. 2018.
\newblock Deep contextualized word representations.
\newblock In \emph{NAACL}.

\bibitem[{Pfeiffer et~al.(2021)Pfeiffer, Kamath, R{\"u}ckl{\'e}, Cho, and Gurevych}]{pfeiffer2020adapterfusion}
Jonas Pfeiffer, Aishwarya Kamath, Andreas R{\"u}ckl{\'e}, Kyunghyun Cho, and Iryna Gurevych. 2021.
\newblock Adapterfusion: Non-destructive task composition for transfer learning.
\newblock In \emph{EACL}.

\bibitem[{Pfeiffer et~al.(2020)Pfeiffer, Vuli{\'c}, Gurevych, and Ruder}]{pfeiffer-etal-2020-mad}
Jonas Pfeiffer, Ivan Vuli{\'c}, Iryna Gurevych, and Sebastian Ruder. 2020.
\newblock {MAD-X}: {A}n {A}dapter-{B}ased {F}ramework for {M}ulti-{T}ask {C}ross-{L}ingual {T}ransfer.
\newblock In \emph{EMNLP}.

\bibitem[{Rajpurkar et~al.(2016)Rajpurkar, Zhang, Lopyrev, and Liang}]{rajpurkar2016squad}
Pranav Rajpurkar, Jian Zhang, Konstantin Lopyrev, and Percy Liang. 2016.
\newblock \href {https://aclanthology.org/D16-1264} {{SQ}u{AD}: 100,000+ questions for machine comprehension of text}.
\newblock In \emph{EMNLP}.

\bibitem[{Sanh et~al.(2020)Sanh, Wolf, and Rush}]{sanh2020movement}
Victor Sanh, Thomas Wolf, and Alexander Rush. 2020.
\newblock Movement pruning: Adaptive sparsity by fine-tuning.
\newblock In \emph{NeurIPS}.

\bibitem[{Socher et~al.(2013)Socher, Perelygin, Wu, Chuang, Manning, Ng, and Potts}]{socher2013recursive}
Richard Socher, Alex Perelygin, Jean Wu, Jason Chuang, Christopher~D Manning, Andrew~Y Ng, and Christopher Potts. 2013.
\newblock Recursive deep models for semantic compositionality over a sentiment treebank.
\newblock In \emph{EMNLP}.

\bibitem[{Tanaka et~al.(2020)Tanaka, Kunin, Yamins, and Ganguli}]{tanaka2020pruning}
Hidenori Tanaka, Daniel Kunin, Daniel~L Yamins, and Surya Ganguli. 2020.
\newblock Pruning neural networks without any data by iteratively conserving synaptic flow.
\newblock In \emph{NeurIPS}.

\bibitem[{Wang et~al.(2020)Wang, Zhang, and Grosse}]{wang2020picking}
Chaoqi Wang, Guodong Zhang, and Roger Grosse. 2020.
\newblock Picking winning tickets before training by preserving gradient flow.
\newblock In \emph{ICLR}.

\bibitem[{Wang et~al.(2021{\natexlab{a}})Wang, Tang, Duan, Wei, Huang, Cao, Jiang, Zhou et~al.}]{wang2020k}
Ruize Wang, Duyu Tang, Nan Duan, Zhongyu Wei, Xuanjing Huang, Guihong Cao, Daxin Jiang, Ming Zhou, et~al. 2021{\natexlab{a}}.
\newblock K-adapter: Infusing knowledge into pre-trained models with adapters.
\newblock In \emph{ACL}.

\bibitem[{Wang et~al.(2021{\natexlab{b}})Wang, Tsvetkov, Ruder, and Neubig}]{wang-etal-2021-efficient-test}
Xinyi Wang, Yulia Tsvetkov, Sebastian Ruder, and Graham Neubig. 2021{\natexlab{b}}.
\newblock Efficient test time adapter ensembling for low-resource language varieties.
\newblock In \emph{EMNLP}.

\bibitem[{Wang et~al.(2022)Wang, Agarwal, Mukherjee, Liu, Gao, Awadallah, and Gao}]{wang-etal-2022-adamix}
Yaqing Wang, Sahaj Agarwal, Subhabrata Mukherjee, Xiaodong Liu, Jing Gao, Ahmed~Hassan Awadallah, and Jianfeng Gao. 2022.
\newblock {A}da{M}ix: Mixture-of-adaptations for parameter-efficient model tuning.
\newblock In \emph{EMNLP}.

\bibitem[{Williams et~al.(2018)Williams, Nangia, and Bowman}]{williams2017broad}
Adina Williams, Nikita Nangia, and Samuel Bowman. 2018.
\newblock \href {https://aclanthology.org/N18-1101} {A broad-coverage challenge corpus for sentence understanding through inference}.
\newblock In \emph{NAACL}.

\end{thebibliography}
\bibliographystyle{acl_natbib}

\newpage
\appendix
\section{Appendix}
\label{sec:appendix}

\begin{table}[htb]
\centering
\small
\begin{tabular}{ccccc}
\toprule
     \textbf{Dataset} & \textbf{mnli} & \textbf{mrpc} & \textbf{sst2} & \textbf{rte} \\
     \midrule
     \textbf{Train} & 392,702 & 3,668 & 67,349 & 2,490   \\ 
     \textbf{Test} & 9,815 & 408 & 872 & 277 \\
     \midrule
     \textbf{Dataset} & \textbf{qnli}  & \textbf{qqp} & \textbf{rct} & \textbf{ag}  \\
     \textbf{Train} & 104,743 & 363,846 & 178,882 &  120,000 \\
     \textbf{Test} & 5,463  & 40,430 & 30,135 &  7,600  \\
     \midrule
    \textbf{authorship} & \textbf{financial} &\textbf{imdb} & \textbf{tweets} &  \textbf{wiki}\\
      2,743 & 4,846 & 22,500 & 31,962  & 127,656 \\
      686 & 484 & 2,500 & 3,196 & 63,978 \\
 \bottomrule
\end{tabular}
\caption{Number of instances for each dataset divided by training and test set.} 
\label{data_stat}  
\end{table}

\begin{figure}[htb]
    \centering
    \includegraphics[width=0.475\textwidth]{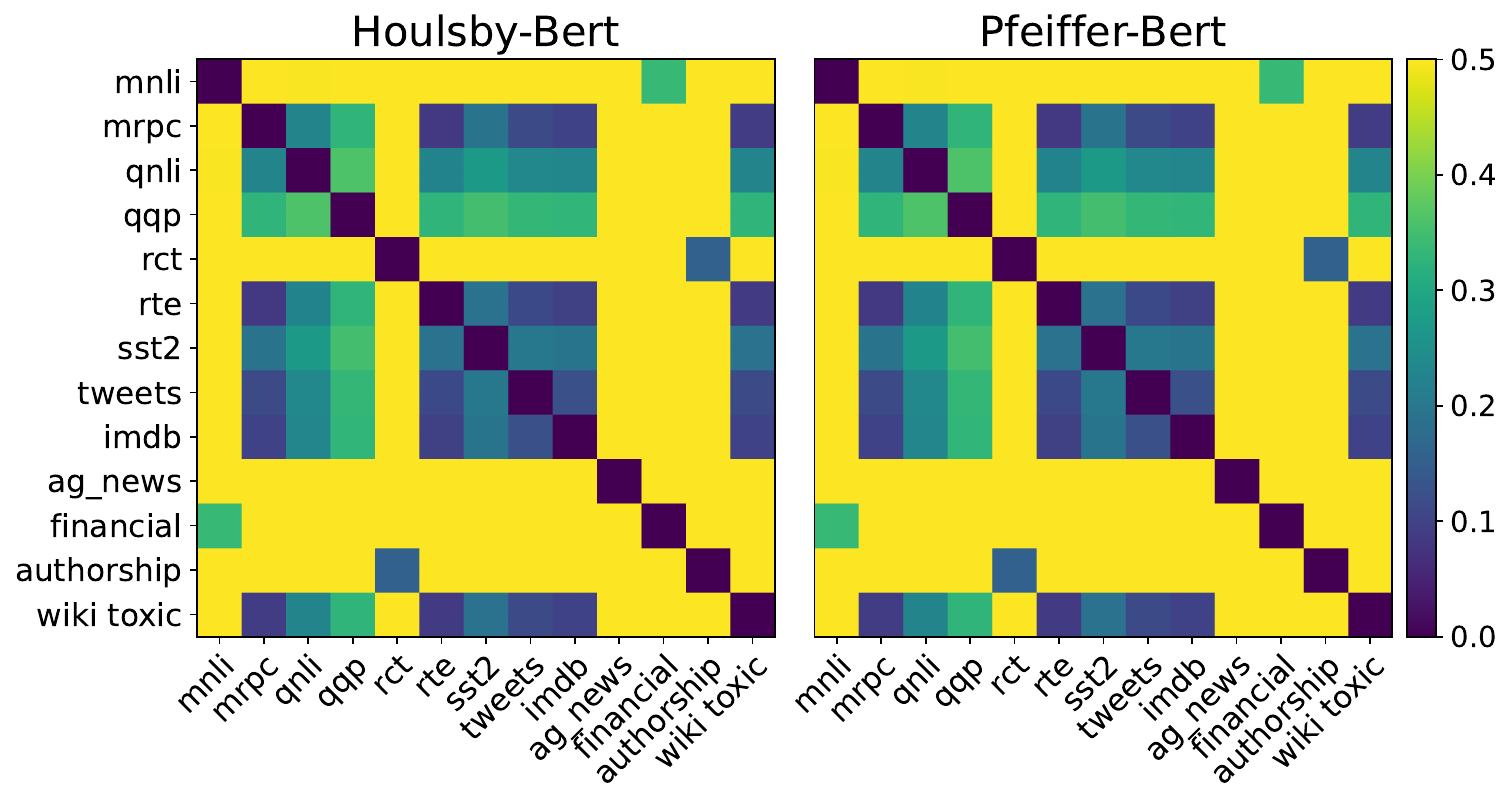}
    \caption{Fraction on differences of adapter weight direction.}
    \label{weight_different_bert}
\end{figure}

\begin{table*}[t]
\setlength{\belowcaptionskip}{-0.1cm}
\small
\centering
\begin{tabular}{cccc}
\toprule
     \textbf{Data Source} & \textbf{\begin{tabular}[c]{@{}c@{}}Average \\ Document Length\end{tabular}} & \textbf{\begin{tabular}[c]{@{}c@{}}Average \\ Sentence Length\end{tabular}} & \textbf{\begin{tabular}[c]{@{}c@{}}Average \\ $\#$ Sentences per Document\end{tabular}} \\
     \midrule
     \textbf{MNLI} & 15.1  & 14.7 & 1.0   \\
     \textbf{MRPC} & 21.9 & 21.1  &  1.0  \\
     \textbf{QNLI} & 18.2  & 18.0 &  1.0  \\
     \textbf{QQP} & 11.1 & 9.9 &  1.2  \\
     \textbf{RTE} & 26.2 & 18.1 &  1.4  \\
     \textbf{SST} & 10.4 & 10.4 &  1.0  \\
 \bottomrule
    \textbf{RCT} & 26.5 & 26.3 &  1.0  \\
    \textbf{Ag-news} & 38.4 & 29.1 &  1.3  \\
     \textbf{Authorship} & 1038.6 & 20.2 &  51.3  \\
     \textbf{Financial} & 23.1  & 22.8 & 1.0   \\
     \textbf{IMDB} & 233.8 & 21.6 &  10.8  \\
     \textbf{Tweets} & 15.9  & 9.6 &  1.6  \\
     \textbf{Wiki-toxic} & 67.8 & 15.4 & 4.4   \\
     \hline
\end{tabular}
\caption{Length statistics.} 
\label{tab:length}
\end{table*}

\begin{algorithm}[ht!]
\footnotesize
\caption{Fraction of weight sign difference (FSD).}
\label{weight_sign_differnce_algorithm}
\textit{\ul{\textbf{Input:}}} two adapters with similar architecture learned from two domain $\theta_A$, $\theta_B$ \\
\textit{\ul{\textbf{Output:}}} Fraction $f$ of weight sign difference \\
\begin{algorithmic}[1]
\State Compute total parameters $s$ in each adapter ($s = s_A = s_B$)
\State {$C \gets \{\}$}
\State For every layer in adapter $\theta_A, \theta_B$, compute \\ element-wise product for each layer.
\State {\bfseries for} $k, v$ {\bfseries in} $\theta_A.items()$ {\bfseries do:}
\State \ \ \ $C[k]$ = $\operatorname{mul}(\theta_A[k],\theta_B[k])$
\State \textit{Count total of numbers which value is \\ smaller than $0$.}
\State {\bfseries for} $k, v$ {\bfseries in} $C.items()$ {\bfseries do:}
\State \ \ \ $counter \operatorname{+=} \operatorname{sum}(value < 0)$
\State $f = counter/ s$
\State {\bfseries return} $f$
\end{algorithmic}
\end{algorithm}

\subsection{Datasets.}
\label{diverse_knowledge_dataset}
\paragraph{Diverse Knowledge Datasets.}
To simulate knowledge diversity, we gather a total of 13 distinct and diverse \textit{domain-specific} datasets or classification tasks for evaluation. They are MNLI~\cite{williams2017broad}, QNLI~\cite{rajpurkar2016squad}, RTE~\cite{bentivogli2009fifth}, MRPC~\cite{dolan2005automatically}, QQP~\cite{iyer2017first} and SST2~\cite{socher2013recursive} from the \textit{GLUE domain} corpus. PubMed-20K RCT dataset~\cite{dernoncourt-lee-2017-pubmed} from \textit{Biology domain} for sentence classification.
IMDB dataset from a \textit{Movie Review domain}. Ag News, Financial~\cite{malo2014good} and Guardian Authorship~\cite{altakrori2021topic} are \textit{News domain} datasets across World, Sports, Business, Science/Technology, and Financial topics. Wiki Toxic \footnote{https://www.kaggle.com/competitions/jigsaw-toxic-comment-classification-challenge/} and Tweets Hate Speech are two \textit{Informal text domain} for toxicity detection.

\paragraph{Linguistic Statistic.}
Table \ref{tab:length} shows detailed statistics such as number of documents, average document length, and sentence lengths.

\subsection{Topic distribution of training datasets}
\label{topic_distribtuion_lda}
Tables from \ref{lda_mnli} to \ref{lda_wiki} show 10 topics and corresponding important words which are exacted from LDA for each training dataset. Notably, Ag News and SST2 have high cosine similarity but observe a large difference in terms of topic distribution compared to other domains (Fig.~\ref{fig:cosine_similarity}). Therefore, each statistical mechanism like cosine similarity or topic distribution only reflects one aspect of data distribution and may show inconsistencies with each other.

\begin{table*}[htb]
\caption{Topic distribution on MNLI dataset}
\label{lda_mnli}
\small
\centering
\resizebox{2\columnwidth}{!}{%
\begin{tabular}{|c|c|}
\hline
\textbf{\#Topic} & \textbf{MNLI}                                                                                                             \\ \hline
1                & well, time, got, take, one, much, day, something, ive, even, way, long, little, make, back                                \\ \hline
2                & kind, system, though, come, went, well, today, view, \redbold{church}, including, \redbold{president}, seems, across, run, \redbold{policy}             \\ \hline
3                & say, get, \redbold{cost}, guess, were, \redbold{business}, car, local, whole, north, rather, getting, question, \redbold{technology}, capital           \\ \hline
4                & service, state, world, get, big, pretty, give, war, yes, standard, real, here, came, call                                 \\ \hline
5                & probably, high, thought, however, set, hand, enough, said, since, type, jon, yet, and, service                            \\ \hline
6                & could, mean, around, part, another, change, percent, made, course, life, book, fact, name, room                           \\ \hline
7                & \redbold{government}, program, federal, information, country, problem, le, new, national, may, number, agency, report, \redbold{organization} \\ \hline
8                & year, two, house, case, old, three, town, street, century, one, city, study, man, four, different                         \\ \hline
9                & know, like, think, thats, right, really, people, thing, good, go, one, lot, going                                         \\ \hline
10               & yeah, work, \redbold{legal}, \redbold{rule}, last, year, he, american, small, home, \redbold{company}, act, group, \redbold{analysis}, public                     \\ \hline
\end{tabular}}
\end{table*}

\begin{table*}[t]
\caption{Topic distribution on MRPC dataset}
\label{lda_mrpc}
\small
\centering
\resizebox{2\columnwidth}{!}{%
\begin{tabular}{|c|c|}
\hline
\textbf{\#Topic} & \textbf{MRPC}                                                                                                      \\ \hline
1                & said, \redbold{court}, company, would, official, \redbold{statement}, \redbold{decision}, made, state, appeal, two, board                        \\ \hline
2                & said, year, people, \redbold{president}, program, time, million, two, last, house, official, weapon                      \\ \hline
3                & said, million, would, state, period, compared, men, get, democratic, plan, company, united, also, could            \\ \hline
4                & percent, \redbold{share}, \redbold{cent}, million, \redbold{stock}, point, \redbold{nasdaq}, billion, new, index, \redbold{trading}, rose, per, year                 \\ \hline
5                & said, also, state, iraq, center, united, \redbold{attack}, hospital, \redbold{killed}, \redbold{war}, three, american, people                    \\ \hline
6                & said, two, home, \redbold{police}, told, state, friday, last, year, federal, company, yesterday, national                    \\ \hline
7                & standard, poor, index, chief, point, said, percent, \redbold{justice}, one, spx, broader, executive, three                   \\ \hline
8                & said, analyst, expected, street, many, suit, call, yesterday, angeles, wall, los, research, one, change, according \\ \hline
9                & case, said, \redbold{court}, filed, death, also, charged, \redbold{lawsuit}, charge, state, found, reported, office, cancer            \\ \hline
10               & said, would, \redbold{server}, window, \redbold{network}, one, new, \redbold{microsoft}, also, taken, people, \redbold{company}                          \\ \hline
\end{tabular}}
\end{table*}

\begin{table*}[t]
\caption{Topic distribution on QNLI dataset}
\label{lda_qnli}
\small
\centering
\resizebox{2\columnwidth}{!}{%
\begin{tabular}{|c|c|}
\hline
\textbf{\#Topic} & \textbf{QNLI}                                                                                                                    \\ \hline
1                & city, american, south, large, west, season, de, roman, service, art, london, first, located, street, new                          \\ \hline
2                & state, united, new, including, \redbold{people}, \redbold{city}, national, million, \redbold{school}, north, \redbold{government}, \redbold{army}, many, within, building           \\ \hline
3                & also, system, later, early, used, based, part, control, four, use, death, official, known, act, called                            \\ \hline
4                & group, language, east, among, found, common, company, india, federal, movement, population, early, included, production, range    \\ \hline
5                & the, \redbold{church}, term, example, \redbold{university}, \redbold{greek}, \redbold{german}, like, english, specie, \redbold{god}, word, per, old, one                            \\ \hline
6                & form, although, following, \redbold{law}, central, \redbold{rule}, culture, without, often, modern, \redbold{territory}, \redbold{society}, treaty, considered, \redbold{christian} \\ \hline
7                & war, world, british, life, development, empire, first, region, community, year, france, though, time, set, began                  \\ \hline
8                & well, three, include, place, power, party, league, may, needed, right, one, political, club, a, event                             \\ \hline
9                & became, first, time, john, film, president, number, year, french, one, day, land, america, process, le                            \\ \hline
10               & century, music, around, house, home, period, age, record, late, established, several, standard, time, world, river                \\ \hline
\end{tabular}}
\end{table*}

\begin{table*}[t]
\caption{Topic distribution on QQP dataset}
\label{lda_qqp}
\small
\centering
\resizebox{2\columnwidth}{!}{%
\begin{tabular}{|c|c|}
\hline
\textbf{\#Topic} & \textbf{QQP}                                                                                                   \\ \hline
1                & like, become, feel, get, job, movie, good, student, want, engineering, girl, website, sex, study, go           \\ \hline
2                & best, way, difference, \redbold{learn}, whats, money, make, online, \redbold{book}, india, buy, start, good, \redbold{language}, \redbold{programming} \\ \hline
3                & much, best, time, weight, year, lose, old, place, month, day, iphone, \redbold{read}, possible, class                \\ \hline
4                & thing, day, business, get, first, going, example, one, prepare, video, woman, word, men                  \\ \hline
5                & work, note, india, indian, ever, computer, black, r, science, you, help, rupee, different           \\ \hline
6                & would, life, \redbold{trump}, \redbold{world}, \redbold{country}, new, donald, \redbold{war}, india, win, happen, \redbold{president}, clinton, hillary       \\ \hline
7                & get, \redbold{friend}, used, long, why, bad, back, see, take, cant, good, facebook, system, \redbold{relationship}, \redbold{person}         \\ \hline
8                & someone, love, english, one, know, improve, account, people, get, instagram, tell, average, hair, password \\ \hline
9                & mean, app, song, name, android, give, bank, right, what, company, india, working, get, now, create             \\ \hline
10               & people, quora, question, think, do, me, answer, google, stop, use, state, get, many, live               \\ \hline
\end{tabular}}
\end{table*}

\begin{table*}[t]
\caption{Topic distribution on RTE dataset}
\label{lda_rte}
\small
\centering
\resizebox{2\columnwidth}{!}{%
\begin{tabular}{|c|c|}
\hline
\textbf{\#Topic} & \textbf{RTE}                                                                                                          \\ \hline
1                & year, bank, world, ago, police, place, human, people, said, man, problem, game, many, took, explosion                 \\ \hline
2                & people, \redbold{attack}, california, \redbold{killed}, life, united, day, lost, air, one, space, \redbold{injured}, national, capital, said        \\ \hline
3                & oil, said, \redbold{nuclear}, company, new, president, iran, million, \redbold{military}, john, un, country, bush, price               \\ \hline
4                & said, world, state, united, minister, country, million, people, nobel, south, peace, \redbold{war}, trade, prize, mexico        \\ \hline
5                & woman, corp, parliament, case, confirmed, said, rabies, represented, cause, poorly, fire, president, \redbold{police}, loss \\ \hline
6                & year, new, said, one, would, died, university, show, company, family, first, service, since, country, home            \\ \hline
7                & state, iraq, said, bush, bomb, found, used, water, home, \redbold{killed}, caused, \redbold{damage}, one, \redbold{police}                       \\ \hline
8                & party, \redbold{police}, president, new, two, officer, name, drug, state, prime, people, \redbold{minister}, last, year, \redbold{democratic}       \\ \hline
9                & new, said, \redbold{government}, year, iraq, would, york, official, today, baghdad, also, euro, announced, percent, minister    \\ \hline
10               & said, year, leader, new, sanfrancisco, work, justice, two, president, \redbold{government}, end, free, guerrilla           \\ \hline
\end{tabular}}
\end{table*}

\begin{table*}[t]
\caption{Topic distribution on SST2 dataset}
\label{lda_sst2}
\small
\centering
\resizebox{2\columnwidth}{!}{%
\begin{tabular}{|c|c|}
\hline
\textbf{\#Topic} & \textbf{SST2}                                                                                                         \\ \hline
1                & \redbold{film}, really, enough, movie, something, make, interesting, many, like, subject, intelligent, laugh, short             \\ \hline
2                & \redbold{movie}, bad, \redbold{film}, better, great, fun, one, look, director, \redbold{story}, ultimately, smart, \redbold{cinema}, put                      \\ \hline
3                & \redbold{performance}, funny, way, moment, film, cast, another, screen, yet, big, work, perfect, made                           \\ \hline
4                & new, material, ve, \redbold{movie}, rather, \redbold{film}, special, seen, minute, enjoyable, might, offer, story, effect                 \\ \hline
5                & \redbold{comedy}, \redbold{drama}, \redbold{thriller}, \redbold{romantic}, \redbold{documentary}, actor, moving, clever, \redbold{funny}, sometimes, pleasure, often, movie, film \\ \hline
6                & work, \redbold{film}, \redbold{movie}, hard, well, keep, filmmaker, ever, life, original, sense, dull, quite, could                       \\ \hline
7                & like, feel, \redbold{movie}, much, people, \redbold{film}, make, see, get, \redbold{character}, one, thing                                          \\ \hline
8                & good, real, \redbold{film}, worth, fascinating, make, time, lack, bit, amusing, humor, tale, pretty, run                        \\ \hline
9                & character, one, best, \redbold{film}, \redbold{movie}, \redbold{story}, far, compelling, two, every, year, picture, little                          \\ \hline
10               & love, audience, \redbold{film}, story, character, seems, entertainment, way, powerful, care, take, one, \redbold{movie}, spirit           \\ \hline
\end{tabular}}
\end{table*}

\begin{table*}[t]
\caption{Topic distribution on RCT dataset}
\label{lda_rct}
\small
\centering
\resizebox{2\columnwidth}{!}{%
\begin{tabular}{|c|c|}
\hline
\textbf{\#Topic} & \textbf{RCT}                                                                                                                                        \\ \hline
1                & group, patient, week, randomized, study, received, control, year, mg, randomly, placebo, day                                               \\ \hline
2                & \redbold{patient}, session, visit, cohort, failure, lesion, myocardial, \redbold{hospital}, twice,\redbold{death}, \redbold{heart}, infarction                          \\ \hline
3                & analysis, using, data, model, used, test, sample, analyzed, regression, characteristic, time, collected, cell, method, performed                    \\ \hline
4                & outcome, primary, month, \redbold{patient}, baseline, measure, score, secondary, \redbold{treatment}, scale, assessed, \redbold{symptom}, week, followup                       \\ \hline
5                & risk, associated, level, \redbold{weight}, factor, effect, \redbold{disease}, body, increased, \redbold{diabetes}, \redbold{insulin}, high, glucose, change, activity                       \\ \hline
6                & study, \redbold{patient}, \redbold{treatment}, effect, \redbold{therapy}, efficacy, may, effective, result, evaluate, safety, weather, clinical, outcome, intervention            \\ \hline
7                & group, difference, significant, significantly, compared, control, treatment, score, higher, lower, time, observed, rate                     \\ \hline
8                & trial, study, randomized, intervention, care, \redbold{health}, controlled, \redbold{clinical}, quality, life, conducted, prospective, effectiveness, child, number     \\ \hline
9                & \redbold{patient}, event, \redbold{surgery}, adverse, postoperative, complication, procedure, \redbold{pain}, undergoing, surgical, rate, incidence, common, \redbold{infection}, injection \\ \hline
10               & mean, respectively, ratio, patient, group, median, versus, interval, year, day, month                                                 \\ \hline
\end{tabular}}
\end{table*}

\begin{table*}[t]
\caption{Topic distribution on Tweets dataset}
\label{lda_tweet}
\small
\centering
\resizebox{2\columnwidth}{!}{%
\begin{tabular}{|c|c|}
\hline
\textbf{\#Topic} & \textbf{Tweets}                                                                                              \\ \hline
1                & new, get, here, music, home,\redbold{ cool}, playing, free, want, \redbold{fun}, season, shop, update, reason                    \\ \hline
2                & day, one, night, time, good, week, last, never, first, get, year, got, lot, today                            \\ \hline
3                & day, father, love, \redbold{happy}, time, weekend, take, friday, dad, fathersday, model                                \\ \hline
4                & want, bull, up, do, help, trump, whatever, direct, dominate, waiting, libtard, yet, sleep, post              \\ \hline
5                & thankful, need, good, positive, orlando, morning, city, tear, news, blessed, friend, dream, bing, yeah, bong \\ \hline
6                & user, amp, day, see, cant, go, like, new, today, one, people, get, wait, make                            \\ \hline
7                & birthday, like, positive, affirmation, happy, baby, amp, god, girl, woman, feel, \redbold{hate}, hot, you           \\ \hline
8                & love, work, life, happy, happiness, make, always, food, quote, smile, wedding, moment, right, feeling, music \\ \hline
9                & healthy, blog, gold, silver, altwaystoheal, forex, healing, grateful, dog, buffalo, peace, really, story     \\ \hline
10               & \redbold{love}, me, \redbold{smile}, summer, beautiful, \redbold{fun}, \redbold{cute}, girl, selfie, friend, sun, instagood, beach, photo            \\ \hline
\end{tabular}}
\end{table*}

\begin{table*}[t]
\caption{Topic distribution on IMDB dataset}
\label{lda_imdb}
\small
\centering
\resizebox{2\columnwidth}{!}{%
\begin{tabular}{|c|c|}
\hline
\textbf{\#Topic} & \textbf{IMDB}                                                                                                                        \\ \hline
1                & story, \redbold{film}, life, \redbold{movie}, character, one, love, time, people, see, way, family, would, well                                          \\ \hline
2                & \redbold{movie}, like, one, \redbold{good}, really, it, \redbold{film}, \redbold{bad}, see, even, time, would, make, get                                                     \\ \hline
3                & get, one, man, the, go, woman, take, back, he, find, there, scene, two, girl                                                         \\ \hline
4                & hamilton, gadget, arkin, scooby, talespin, stallion, smoothly, tenderness, shaggy, gil, inspector, keller, nevada, hopelessness \\ \hline
5                & \redbold{war}, american, documentary, \redbold{soldier}, \redbold{political}, world, german, country, \redbold{history}, america, \redbold{military}, army, hitler            \\ \hline
6                & \redbold{bollywood}, \redbold{indian}, kapoor, khan, akshay, fi, amitabh, ramones, verhoeven, christina, sci, braveheart, kumar, chiller                 \\ \hline
7                & \redbold{film}, one, the, scene, \redbold{character}, story, \redbold{director}, much, plot, well, even, work, time                                          \\ \hline
8                & \redbold{film}, role, \redbold{performance}, great, play, best, good, \redbold{cast}, one, \redbold{actor}, \redbold{comedy}, john                                         \\ \hline
9                & \redbold{show}, \redbold{series}, \redbold{episode}, year, tv, time, great, first, kid, dvd, one, funny, still, watch                                          \\ \hline
10               & match, matthau, luke, \redbold{shakespeare}, neil, bruce, scarface, boxing, \redbold{hamlet}, elvis, branagh, lucas, polanski                            \\ \hline
\end{tabular}}
\end{table*}

\begin{table*}[t]
\small
\centering
\caption{Topic distribution on Ag News dataset}
\label{lda_ag_news}
\resizebox{2\columnwidth}{!}{%
\begin{tabular}{|c|c|}
\hline
\textbf{\#Topic} & \textbf{Ag News}                                                                                                   \\ \hline
1                & palestinian, said, iraqi, \redbold{killed}, iraq, reuters, \redbold{attack}, baghdad, arafat, israeli, \redbold{bomb}, scored, force, city       \\ \hline
2                & \redbold{win}, world, first, point, \redbold{coach}, \redbold{cup}, lead, \redbold{victory}, team, second, no, \redbold{champion}, night, \redbold{final}                      \\ \hline
3                & \redbold{president}, afp, said, \redbold{minister}, \redbold{election}, bush, leader, india, state, reuters, prime, united                       \\ \hline
4                & reuters, \redbold{oil}, \redbold{price}, \redbold{stock}, new, search, \redbold{dollar}, google, market, york, rate, apple, \redbold{share}, record                  \\ \hline
5                & court, \redbold{drug}, say, ap, could, may, new, year, eu, case, said, state, \redbold{scientist}, \redbold{trial}                               \\ \hline
6                & \redbold{space}, \redbold{nasa}, canadian, dec, press, former, nba, williams, winter, houston, monday, arsenal, sunday                 \\ \hline
7                & said, \redbold{company}, \redbold{inc}, \redbold{million}, \redbold{deal}, corp, billion, \redbold{sale}, year, percent, reuters, \redbold{buy}, business                      \\ \hline
8                & \redbold{microsoft}, new, \redbold{software}, internet, service, system, \redbold{computer}, \redbold{technology}, phone, ibm, music, online, web, company \\ \hline
9                & china, police, said, reuters, people, worker, british, government, official, party, japan, group, chinese          \\ \hline
10               & \redbold{game}, new, year, red, one, time, season, first, \redbold{team}, \redbold{series}, last, york                                           \\ \hline
\end{tabular}}
\end{table*}

\begin{table*}[t]
\caption{Topic distribution on Financial dataset}
\label{lda_financial}
\small
\centering
\resizebox{2\columnwidth}{!}{%
\begin{tabular}{|c|c|}
\hline
\textbf{\#Topic} & \textbf{Financial}                                                                                                              \\ \hline
1                & company, finnish, new, plant, finland, construction, order, line, \redbold{contract}, service, unit, production, \redbold{investment}         \\ \hline
2                & company, \redbold{share}, bank, said, also, capital, start, issue, term, \redbold{financial}, price, \redbold{business}, executive, dividend                  \\ \hline
3                & eur, \redbold{profit}, sale, \redbold{net}, operating, million, period, quarter, compared, \redbold{loss}, year                                               \\ \hline
4                & finnish, said, today, million, company, first, helsinki, year                                                                   \\ \hline
5                & \redbold{company}, mobile, said, phone, nokia, solution, business, pretax, finland, network, \redbold{product}, group, store, \redbold{customer}              \\ \hline
6                & \redbold{market}, board, option, company, \redbold{share}, \redbold{stock}, director, member, concerning, meeting, general, \redbold{bank}, flow, chairman              \\ \hline
7                & \redbold{share}, company, group, lower, helsinki, \redbold{stock}, president, capital, holding, new, right                                          \\ \hline
8                & service, finland, customer, corporation, company, electronics, solution, industry, \redbold{business}, helsinki, ltd, group \\ \hline
9                & company, expected, sale, said, people, production, paper, year, finland, plant, cut, staff, expects                             \\ \hline
10               & euro, service, company, item, nokia, excluding, technology, \redbold{business}, mobile, device, \redbold{market}, \redbold{product}                           \\ \hline
\end{tabular}}
\end{table*}

\begin{table*}[t]
\caption{Topic distribution on Authorship dataset}
\label{lda_authorship}
\small
\centering
\resizebox{2\columnwidth}{!}{%
\begin{tabular}{|c|c|}
\hline
\textbf{\#Topic} & \textbf{Authorship}                                                                                \\ \hline
1                & one, would, may, people, year, even, could, time, last, minister, public, police, many, blair, say \\ \hline
2                & one, would, war, farmer, even, new, blair, bush, could, need, time, iraq, much, week               \\ \hline
3                & labour, new, people, \redbold{government}, tax, year, time, even, public, brown, blair, party, money         \\ \hline
4                & would, one, \redbold{government}, new, world, year, \redbold{labour}, much, state, blair, last, british                \\ \hline
5                & new, public, government, \redbold{labour}, people, year, one, would, may, way, time, make, right, life, need \\ \hline
6                & people, time, public, said, even, government, lord, like, party, make, day                         \\ \hline
7                & one, bush, american, world, year, right, war, child, people, british, state, new                   \\ \hline
8                & people, one, child, like, time, family, get, year, burrell, may, still, even, much                 \\ \hline
9                & would, one, blair, bush, war, nuclear, even, it, new, make, could, weapon, people, party           \\ \hline
10               & would, one, year, people, could, even, royal, like, woman, time, war, right, iraq                  \\ \hline
\end{tabular}}
\end{table*}

\begin{table*}[]
\caption{Topic distribution on Wiki Toxic dataset}
\label{lda_wiki}
\small
\centering
\resizebox{2\columnwidth}{!}{%
\begin{tabular}{|c|c|}
\hline
\textbf{\#Topic} & \textbf{Wiki Toxic}                                                                                                      \\ \hline
1                & page, talk, edit, please, user, edits, wikipedia, editor, comment, \redbold{block}, \redbold{blocked}, editing, discussion, thanks, stop     \\ \hline
2                & image, use, you, copyright, page, fair, picture, please, medium, wikipedia, see, template, deleted, file, photo          \\ \hline
3                & article, deletion, deleted, page, please, tag, may, speedy, notable, talk, guideline, subject, wikipedia, criterion, add \\ \hline
4                & \redbold{nigger}, \redbold{hate}, \redbold{bitchfuck}, \redbold{faggot}, lol, class, \redbold{rape}, fat, \redbold{asshole}, mama, \redbold{fucker}, hairy, ha, \redbold{boymamas}                  \\ \hline
5                & like, know, get, people, it, think, you, want, one, time, go, thing, me, really                                      \\ \hline
6                & state, english, country, american, language, people, name, war, city, world, government, history, british, jew, group    \\ \hline
7                & \redbold{fuck}, \redbold{ass}, suck, \redbold{fucking}, \redbold{shit}, u, hi, cunt, school, \redbold{moron}, go, \redbold{bitch}, shut, cock, dick                                   \\ \hline
8                & utc, year, new, game, redirect, song, old                                              \\ \hline
9                & page, wikipedia, talk, help, please, link, welcome, question, article, thank, thanks, like, name, best                \\ \hline
10               & article, one, would, source, also, think, section, fact, see, it, like, point, say, time, reference                      \\ \hline
\end{tabular}}
\end{table*}

\subsection{Hyper-parameter}
\label{sec:hyper}
\noindent \textbf{Training and evaluation datasets.}
To assess performance in out-of-distribution scenarios, we conduct evaluations on a diverse set of 13 datasets covering various topics, ranging from movie reviews, news, authorship, and healthcare, to non-formal language text such as Wiki Toxic and Tweets.
For datasets within the GLUE corpus, we employ training and evaluation datasets to gauge accuracy across different settings. 
In the case of Ag News, Authorship, Financial, IMDB, Tweets, and Wiki-Toxic, we partition the training set into three segments with an 8:1:1 ratio, utilizing them for training, evaluation, and test datasets, respectively. 
This approach ensures a comprehensive evaluation of model performance across a wide spectrum of domains and linguistic styles.
Table \ref{data_stat} shows data statistics on train/test datasets.

\noindent \textbf{Setting on text adversarial attack.}
In this study, we employ two types of attacker methods: \textit{TextFooler} \cite{jin2020bertrobust} and FGSM \cite{goodfellow2015}.

\textit{TextFooler} word-level attacks focus on replacing words within the text with synonyms or contextually similar words. 
By making ostensibly minor alterations to the input text, these attacks can deceive LLMs into producing
incorrect outputs or substantially modifying their predictions. 
We meticulously fine-tune the hyperparameters of TextFooler to obtain more appropriate synonyms.
We set the minimum embedding cosine similarity between a word and its synonyms as 0.8, and the minimum Universal Sentence Encoder similarity is 0.84.

FGSM \cite{goodfellow2015} is a white-box embedding-level attack. 
FGSM uses the Fast Gradient Sign Method to calculate gradients of the model's loss to the input text and generates an adversarial example by perturbing the embedding of input text in the direction that maximizes the loss. 
We choose the magnitude of the perturbation in embedding space as $0.01$ on BERT and RoBERTa models. 

\noindent \textbf{Adapter Configuration.}
We use adapters with a dimension of 64 and 256 using
RoBERTa-large and BERT-base encoders following the setup of \cite{houlsby2019parameter}, \cite{pfeiffer2020adapterfusion}. 
With LoRA, we use rank $r = 4$ following the setup of \cite{hu2021lora}.

\noindent \textbf{Hardware Information.} We evaluate model performance on AMD Ubuntu 22.04.2 with Ryzen Threadripper PRO 5975WX, 1800MHz, Cached 512 KB and 4 $\times$ GPU Nvidia A6000.
\noindent \textbf{Hyper-Parameters.} 
Detailed hyper-parameter configuration for different tasks is presented in Table~\ref{tab:hyper-glue_dev}.

\begin{table*}
\small
	\begin{center}
		\begin{tabular}{@{\hskip1pt}l@{\hskip1pt}|@{\hskip1pt}c@{\hskip1pt}|c@{\hskip1pt}|c@{\hskip1pt}|c@{\hskip1pt}|c@{\hskip1pt}|@{\hskip1pt}c @{\hskip1pt}|@{\hskip1pt}}
			\toprule \bf Task & Learning rate & epoch       &batch size        &warmup                &weight decay  & adapter size   \\ \midrule \bottomrule
	
	\multicolumn{7}{c}{\textbf{BERT\textsubscript{BASE}}}\\ \midrule	
	    MNLI&4e-4&20&32& 0.06 & 0.1 & 256 \\
           MRPC& 4e-4 & 5 & 32 & 0.06 & 0.1 & 256 \\
           QNLI&4e-4&20&32& 0.06 & 0.1 & 256 \\
           QQP&4e-4&20&32& 0.06 & 0.1 & 256 \\
           RCT&4e-4&20&32& 0.06 & 0.1 & 256 \\
           RTE&4e-4&5&32& 0.06 & 0.1 & 256 \\
           SST2&4e-4&10&32& 0.06 & 0.1 & 256 \\
           Tweets&4e-4&5&32& 0.06 & 0.1 & 256 \\
           IMDB&4e-4&5&32& 0.06 & 0.1 & 256 \\
           Ag News&4e-4&20&32& 0.06 & 0.1 & 256 \\
           Financial&4e-4&5&32& 0.06 & 0.1 & 256 \\
           Authorship&4e-4&5&32& 0.06 & 0.1 & 256 \\
           	\midrule
        \multicolumn{7}{c}{\textbf{RoBERTa\textsubscript{LARGE}}}\\ \midrule	
           MNLI&3e-4&20&64&0.6&0.1&64\\
           MRPC&3e-4&5&64&0.6&0.1&64\\
           QNLI&3e-4&20&64&0.6&0.1&64\\
           QQP&3e-4&20&64&0.6&0.1&64\\
           RCT&3e-4&20&64&0.6&0.1&64\\
           RTE&3e-4&5&64&0.6&0.1&64\\
           SST2&3e-4&10&64& 0.6 & 0.1 & 64 \\
           Tweets&3e-4&5&64& 0.6 & 0.1 & 64 \\
           IMDB&3e-4&5&64& 0.6 & 0.1 & 64 \\
           Ag News&3e-4&20&64& 0.6 & 0.1 & 64 \\
           Financial&3e-4&5&64& 0.6 & 0.1 & 64 \\
           Authorship&3e-4&5&64& 0.6 & 0.1 & 64 \\  
               	\bottomrule
        
		\end{tabular}
	\end{center}
	\caption{Hyperparameter configurations for various tasks.}
	\label{tab:hyper-glue_dev}
\end{table*}

\subsection{Model performance when mixing adapters across tasks}
\label{additional_results_mixing_all}
\begin{figure*}[t]
    \centering
    \includegraphics[width=\textwidth]{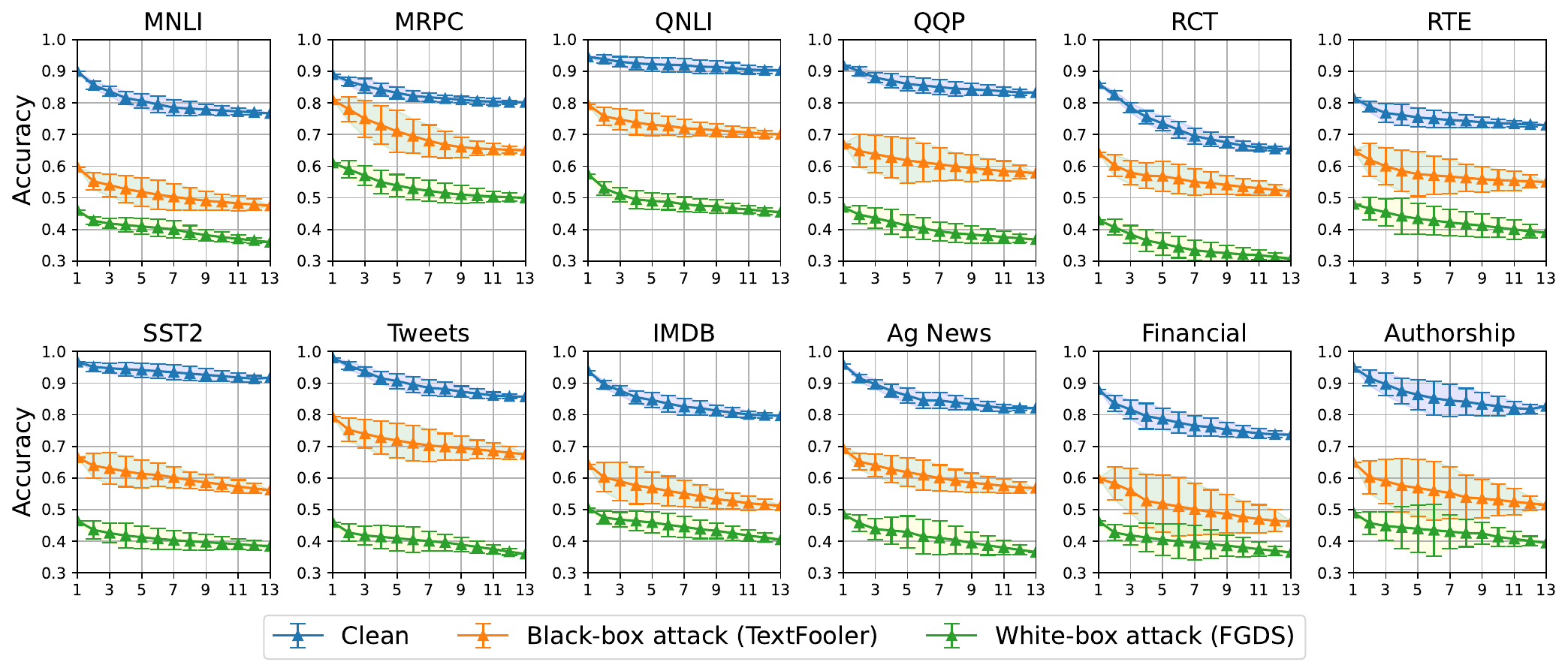}
    \caption{Accuracy of RoBERTa with Houlsby~\cite{houlsby2019parameter} across various distribution datasets. \textit{The x-axis denotes the number of domain adapters to be mixed, ranging from 1 to 13}.}
    \label{roberta_houlsby}
\end{figure*}

\begin{figure*}[t]
    \centering
    \includegraphics[width=\textwidth]{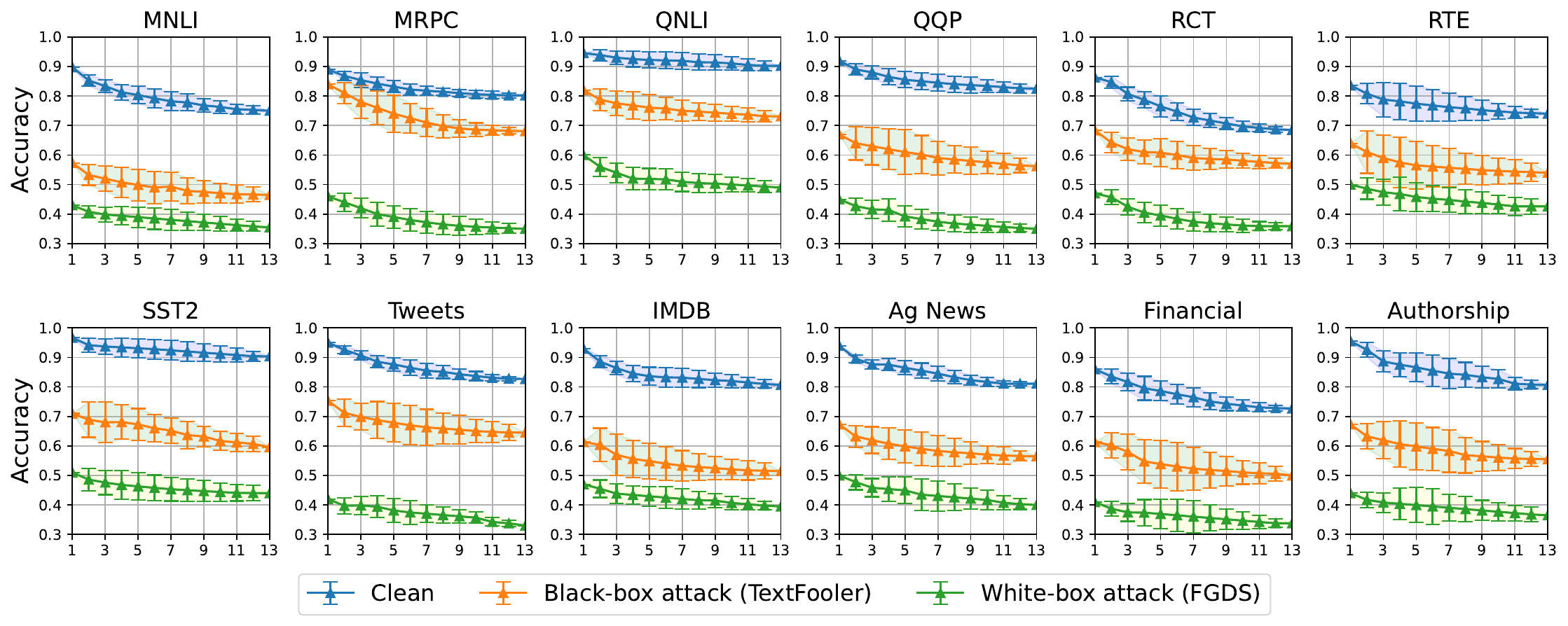}
    \caption{Performance Evaluation of RoBERTa Using the LoRA~\cite{hu2021lora} across Varied Domain Datasets.}
    \label{roberta_lora}
\end{figure*}

\begin{figure*}[t]
    \centering
    \includegraphics[width=\textwidth]{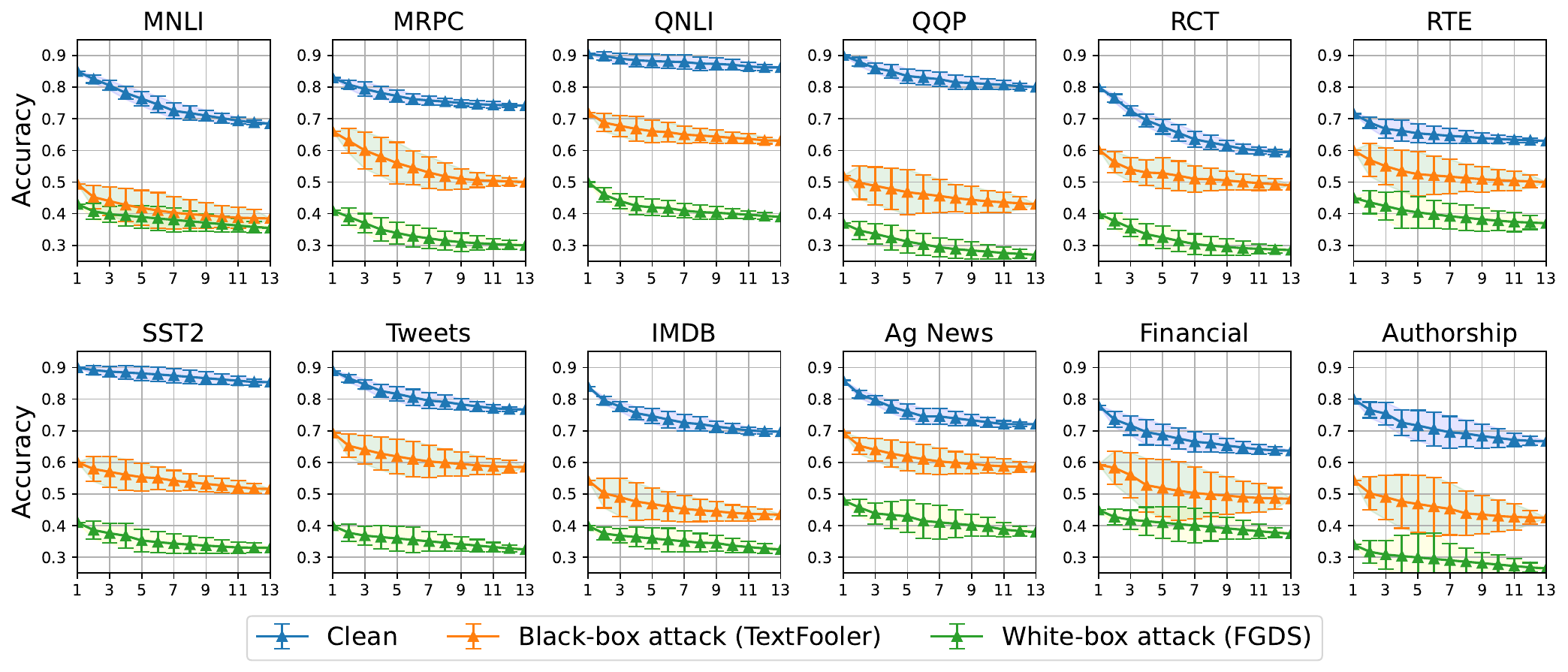}
    \caption{Performance Evaluation of BERT Using the Houlsby~\cite{houlsby2019parameter} across Varied Domain Datasets.}
    \label{bert_houlsby}
\end{figure*}

\begin{figure*}[t]
    \centering
    \includegraphics[width=\textwidth]{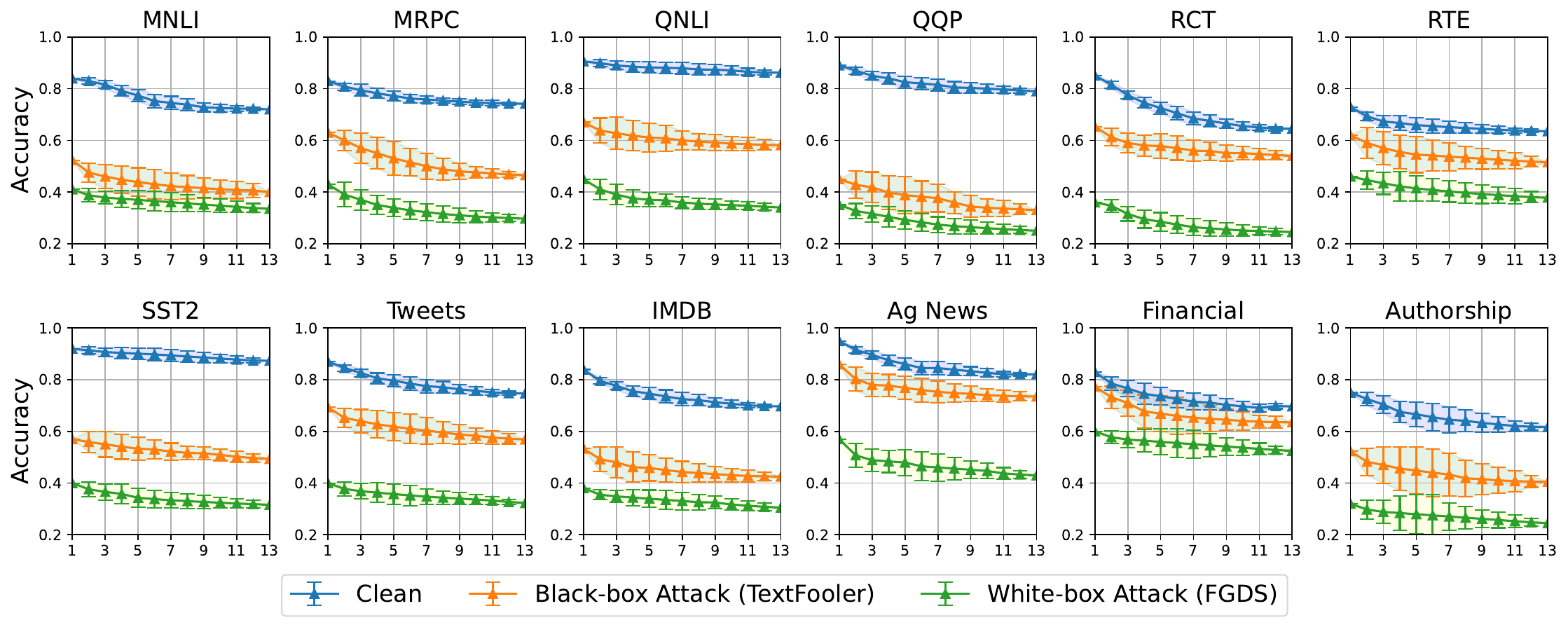}
    \caption{Performance Evaluation of BERT Using the Pfeiffer~\cite{pfeiffer2020adapterfusion} across Varied Domain Datasets.}
    \label{bert_pfeiffer}
\end{figure*}

Tables \ref{roberta_houlsby}, \ref{roberta_lora}, \ref{bert_houlsby} and \ref{bert_pfeiffer} show task accuracy when mixing multiple adapters.

\subsection{Fraction of weight sign difference}
\label{sign_diff_section}
Alg.~\ref{weight_sign_differnce_algorithm} presents a detailed algorithm to compute the FSD.

\subsection{Additional result on weight sign difference}
\label{additional_result_bert_diff}
Fig.~\ref{weight_different_bert} shows the weight sign difference of the adapter and normalizes it by the total number of adapter weights in BERT.

\subsection{Additional results when mixing two adapters}
\label{two_adapter_diff_full}
\begin{figure*}[t]
    \centering
    \includegraphics[width=0.95\textwidth]{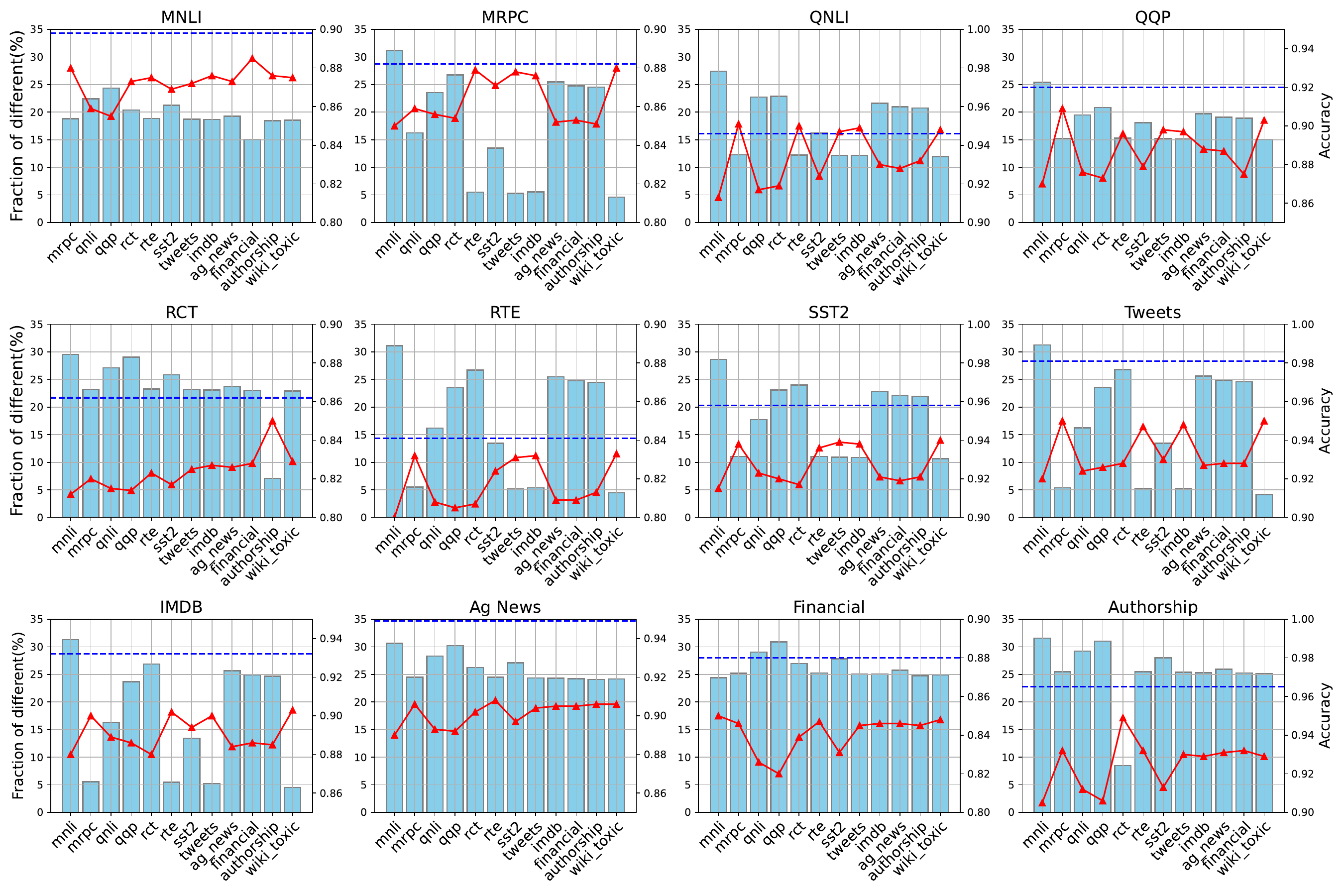}
    \caption{Fraction of weights altering direction during the consolidation of two adapters. The \textcolor{lightskyblue}{sky-blue} bar represents the fraction of weight sign conflicts between two (k=2) adapters (left y-axis).
    The \textit{dashed \textcolor{blue}{blue} line} denotes the accuracy achieved by a standalone adapter trained on a specific task. 
    While the \textit{solid \textcolor{red}{red} line} illustrates the variations in accuracy when merging the adapter with another task's adapter.}
    \label{pfeiffer_roberta_two_adapters_diff_full}
    \vspace{-5pt}
\end{figure*}

\begin{figure*}[t]
    \centering
    \includegraphics[width=0.9\textwidth]{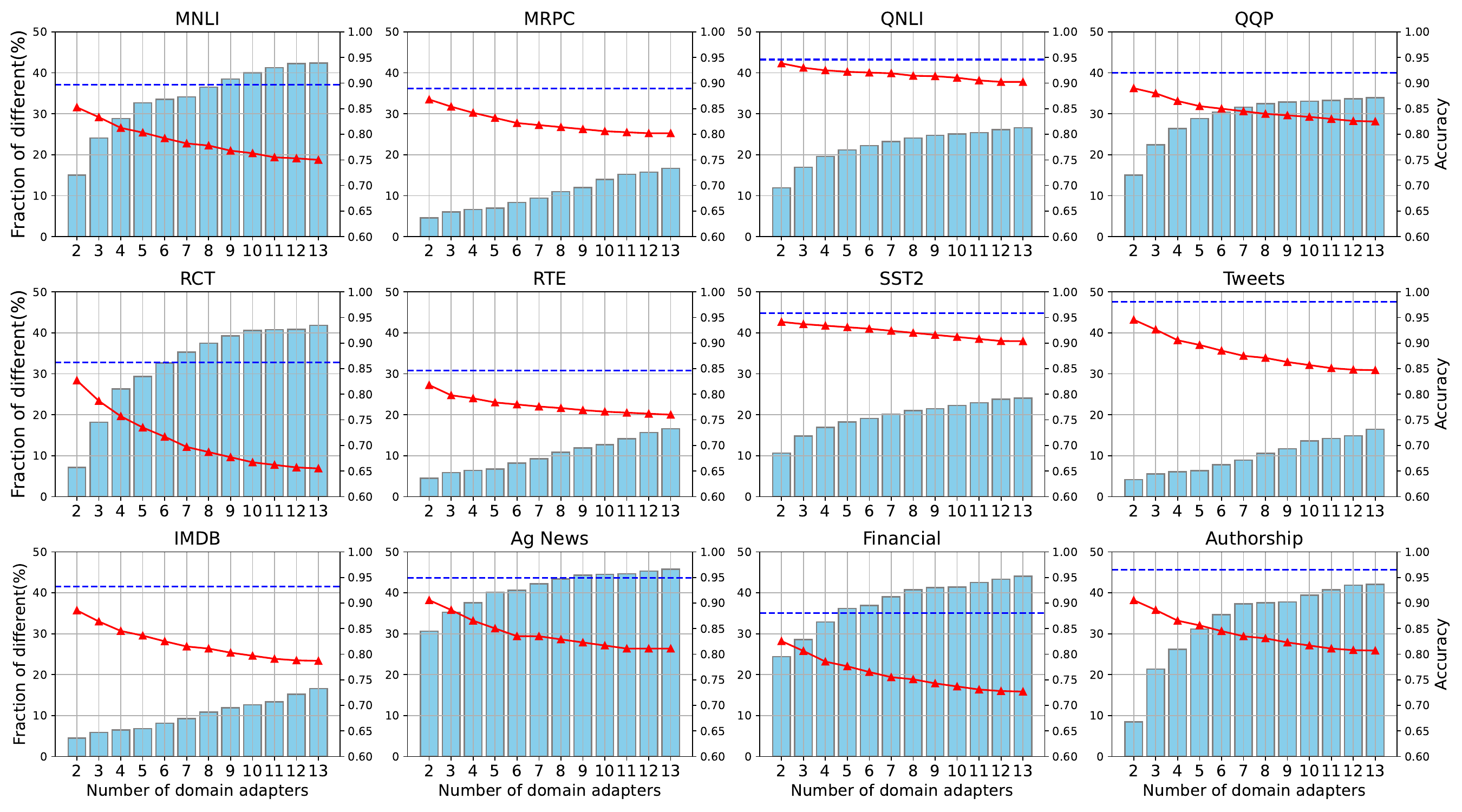}
    \caption{Fraction of weights changing direction during the mixing of multiple adapters, ranging from 2 to 13. 
    The \textcolor{lightskyblue}{sky-blue} bar represents the fraction of weight sign conflicts between k (from 1 to 13) adapters (left y-axis).
    The \textit{dashed \textcolor{blue}{blue}} line corresponds to the accuracy of a single adapter trained on a specific task, while the \textit{solid \textcolor{red}{red} line} depicts the fluctuation in task accuracy resulting from merging the adapter with another task's adapter.}
    \label{pfeiffer_roberta_multi_adapters_diff_full}
    \vspace{-5pt}
\end{figure*}

Fig.~\ref{pfeiffer_roberta_two_adapters_diff_full} and \ref{pfeiffer_roberta_multi_adapters_diff_full} show model generalization when mixing two and multiple adapters across various tasks.

\subsection{Mixing Sparse Adapters with weight sign difference}
\label{mixing_sparse_with_sign_diff}
Alg.~\ref{mixing_pruned_adapter_with_sign_conflict_information} shows details of mixing sparse adapter with sign conflict information.

\begin{algorithm}[tb]
\footnotesize
\caption{Mixing Sparse Adapters with weight sign difference}
\label{mixing_pruned_adapter_with_sign_conflict_information}
\textit{\ul{\textbf{Input:}}} $k$ domain-specific adapters, the FSD matrix  $\mathbf{S}_{k\times k}$, $l$ number of mix adapters.\\

\textit{\ul{\textbf{Output:}}} Average($\mathsf{sparse\_candidates}$) \\
\begin{algorithmic}[1]
\State{$\mathsf{dense\_candidates} \gets \{ \}$}
\State{Compute the average of the difference in weight \\ sign: $average_S = mean(S, axis = 1)$}
\State{Select the top $l$ smallest adapters ($\mathbf{S}_l$) to mix based \\ on the average weight sign difference}
\State ${\mathsf{dense\_candidates} \gets \mathbf{S}_l}$
\State For each adapter, compute the average fraction of \\ weight sign different in each layer with corresponding layers from other adapters.
\State Get the top $m$ layer with the highest fraction of \\ weight sign conflict to prune
\State $\mathsf{sparse\_candidates} \gets \operatorname{Prune}(\mathsf{dense\_candidates})$
\State {\bfseries return} $\operatorname{average}(\operatorname{\mathsf{sparse\_candidates}})$
\end{algorithmic}
\end{algorithm}

\end{document}